\pdfoutput=1
\documentclass[11pt]{article}

\usepackage[]{acl}

\usepackage{times}
\usepackage{latexsym}

\usepackage[T1]{fontenc}
\usepackage[utf8]{inputenc}

\usepackage{microtype}

\usepackage{inconsolata}
\usepackage{graphicx}
\usepackage{algorithm}
\usepackage{algorithmic}
\usepackage{amssymb}% http://ctan.org/pkg/amssymb
\usepackage{pifont}% http://ctan.org/pkg/pifont
\usepackage{multirow}
\usepackage{listings}
\usepackage{newfloat}
\usepackage{amsmath}
\usepackage{xcolor,colortbl}
\usepackage{enumitem}
\definecolor{lightgreen}{rgb}{0.60, 0.78, 0.75}
\definecolor{lightorange}{rgb}{1.0, 0.949, 0.80}
\definecolor{verylightorange}{rgb}{1.0, 0.875, 0.502}
\definecolor{verylightgreen}{rgb}{0.796, 0.886, 0.871}
\title{VEIL: Vetting Extracted Image Labels from In-the-Wild Captions for Weakly-Supervised Object Detection}

\author{Arushi Rai\\
University of Pittsburgh\\
  \texttt{arr159@pitt.edu} \\\And
Adriana Kovashka\\
University of Pittsburgh\\
  \texttt{kovashka@cs.pitt.edu} \\}

\begin{document}
\maketitle
\begin{abstract}
The use of large-scale vision-language datasets is limited for object detection due to the negative impact of label noise on localization. Prior methods have shown how such large-scale datasets can be used for pretraining, which can provide initial signal for localization, but is insufficient without clean bounding-box data for at least some categories. We propose a technique to ``vet'' labels extracted from noisy captions, and use them for weakly-supervised object detection (WSOD), without any bounding boxes. We analyze and annotate the types of label noise in captions in our Caption Label Noise dataset, and train a classifier that predicts if an extracted label is actually present in the image or not. Our classifier generalizes across dataset boundaries and across categories. We compare the classifier to nine baselines on five datasets, and demonstrate that it can improve WSOD without label vetting by 30\% (31.2 to 40.5 mAP when evaluated on PASCAL VOC). See dataset at: \url{https://github.com/arushirai1/CLaNDataset}.
\end{abstract}
\section{Introduction}

Freely available vision-language (VL) data has shown great promise in advancing vision tasks \cite{Radford2021LearningTV,wslimageseccv2018,Jia2021ScalingUV}. 
Unlike smaller, curated vision-language datasets like COCO \cite{Lin2014MicrosoftCC}, 
captions on the web \cite{Ordonez2011Im2TextDI, Desai2021RedCapsWI, Changpinyo2021Conceptual1P} only \textit{partially} describe the corresponding image, and often describe the \textit{context}, which could include
objects that do not appear in the image. We hypothesize this 
poses a greater challenge for weakly-supervised object detection (WSOD) than learning cross-modal representations for image recognition (e.g. as in CLIP). 
WSOD involves learning to localize objects, i.e. predict bounding box coordinates along with the corresponding semantic label, from image-level labels only (i.e. using weaker supervision than the outputs expected at test time). So, noise could compound the challenge of implicitly learning localization.
WSOD has primarily been applied \cite{ye_2019_cap2det, Fang2022DataDD} to smaller, relatively cleaner, paid-for crowdsourced vision-language datasets like COCO \cite{Lin2014MicrosoftCC} and Flickr30K \cite{Young2014Flicker30K}.

\begin{figure}[t]
\centering
    \includegraphics[width=\linewidth]{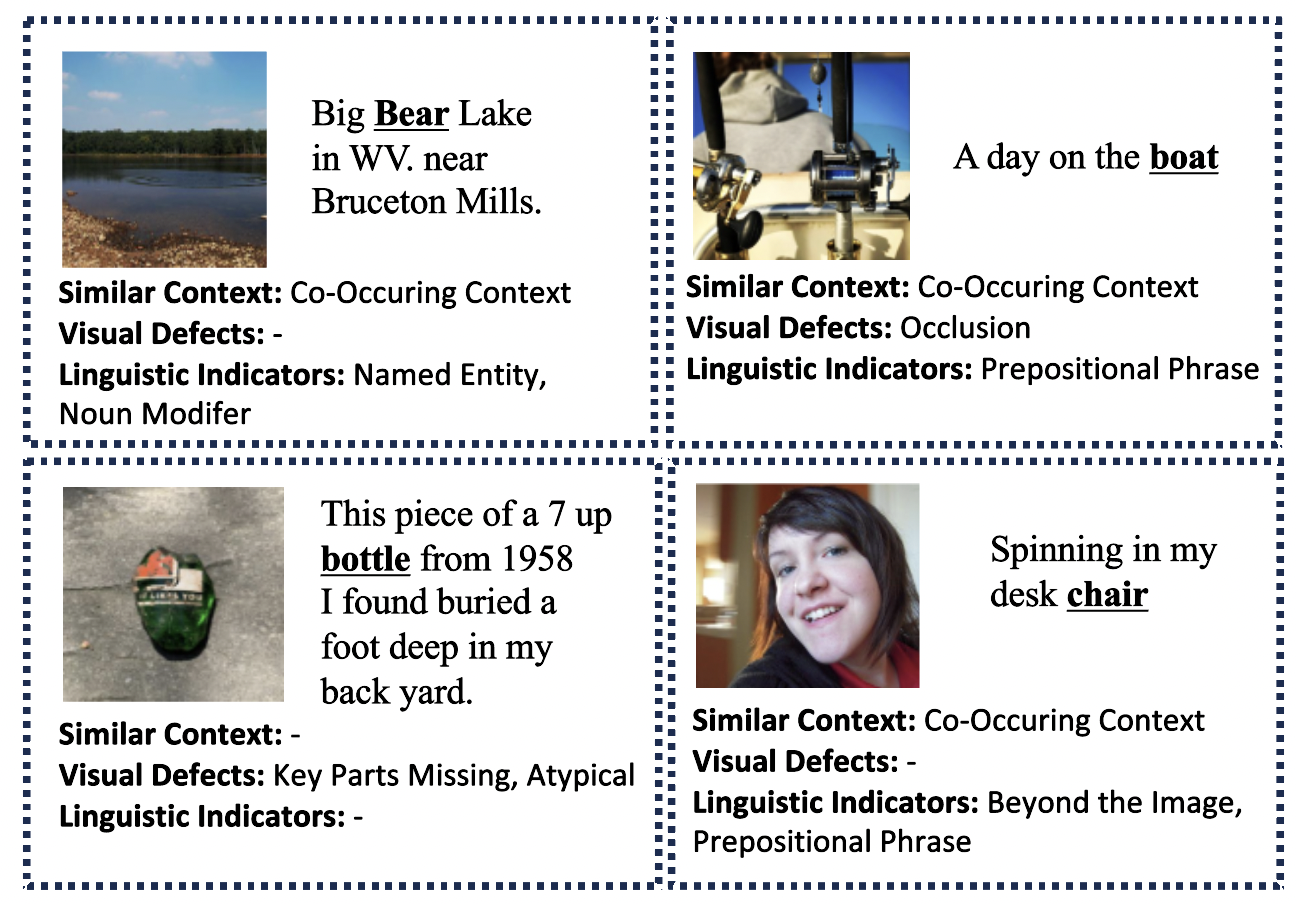}
    % \caption{Extracted labels from captions raise challenges such as missing objects or defects, annotated in our dataset, \textbf{C}aption \textbf{La}bel \textbf{N}oise. \textbf{None of the underlined objects are clearly visible.} We propose a method to detect such noise and compare it to alternatives.}
    \caption{Examples of noisy extracted labels (underlined%in captions
    ) from our \textbf{C}aption \textbf{La}bel \textbf{N}oise dataset. 
    % Each caption contains an underlined object that is not visible in the corresponding image. 
    We categorize types of similar context present instead of the underlined object, as well as types of visual defects and linguistic indicators that are useful for detecting noise.}
    \label{fig:concept}
\end{figure}

We argue that extending WSOD from paid-for captions to large-scale, in-the-wild captions is not trivial. 
% Unlike captions written by 
Annotators write captions that %for the purpose of
faithfully describe an image, however, web captions %on the web 
go beyond a 
% redundant, 
descriptive relationship with their corresponding image.
%in many ways. 
For example, a word can be used literally or metaphorically (``that was a piece of \underline{cake}'') or have multiple senses, of which only one sense 
is relevant to the object detection vocabulary. A caption could also share a story and include context that goes beyond the visual contents of the image; this context could mention an object name within location names or describe occluded or unpictured interactions with objects as shown in Figure \ref{fig:concept}. \textbf{This richness of language is relevant as narration for the image but not as supervision for the precise localization of objects.} On the visual side, user-uploaded content frequently features diverse object presentations, including intriguing atypical objects, hand-drawn objects, or photos taken from within vehicles (``in my \underline{car}'').

We refer to image-level labels extracted from captions, that are incorrect (object not present in the corresponding image), as \textbf{v}isually \textbf{a}bsent \textbf{e}xtracted \textbf{l}abels (VAELs). We show VAELs pose a challenge for weakly-supervised object detection.
 
To cope with this challenge, we propose \textbf{VEIL}, short for \textbf{V}etting \textbf{E}xtracted \textbf{I}mage \textbf{L}abels, to directly learn whether a label is clean or not from \textit{caption context}. 
We first extract potential labels from each caption using substring matching or exact match \cite{Ye_Zhang_Kovashka_Li_Qin_Berent_2019, Fang2022DataDD}. 
We then use a transformer to 
predict whether each extracted label is visually present. We refer to this prediction \emph{task} as extracted label \underline{vetting}.
We bootstrap %image-level 
pseudo-ground-truth visual presence labels for each extracted label or object mention using an ensemble of two pretrained object recognition models \cite{yolov5, zhang2021vinvl}, for a variety of large-scale, noisy datasets: Conceptual Captions \cite{Sharma2018ConceptualCA}, RedCaps \cite{Desai2021RedCapsWI}, and SBUCaps \cite{Ordonez2011Im2TextDI}. While 
these models are trained on COCO and similar datasets, they generalize well to estimating extracted label visual presence on in-the-wild VL datasets; however, their predictions are better used as targets for VEIL, rather than directly for vetting. 
Once we vet the extracted labels, we use them to train a weakly-supervised object detector. 

% We investigate sources of noise across three in-the-wild datasets from diverse sources: a photo-sharing platform, a social media platform, and images with alt-text (typically used for VL pertaining). 
We collect and release the \textbf{C}aption \textbf{La}bel \textbf{N}oise (CLaN) dataset with annotations on object visibility (label noise) and object appearance defects (visual noise such as atypical appearance) over three in-the-wild datasets. 
To support using language context to filter object labels, we annotate linguistic indicators of noise that explain \textit{why} an object is absent from the image but mentioned in the caption. 
%These indicators include describing context outside the image, non-literal use, different word sense, etc. 
% We compare 
Our label vetting method outperforms nine diverse baselines, including standard cross-modal alignment prediction methods (CLIP), adaptive noise reduction methods, pseudo-label prediction, simple rule-based methods, and no vetting. 
% Our method improves upon the baselines both in terms of predicting extracted label visual presence (measured with F1) and producing cleaner training data for object detection. 
This means VEIL produces cleaner WSOD training data which leads to an improvement of +10 mAP over data cleaned using Large Loss Matters \cite{Kim2022LargeLM} and +3 mAP improvement over using CLIP \cite{Radford2021LearningTV} for filtering.
%, when we train on a held-out SBUCaps subset. 
Our findings reveal that naively combining noisy SBUCaps supervision with clean labels from Pascal VOC-07 degrades performance (42.06 mAP) versus using only clean labels (43.48 mAP); however, vetting with VEIL improves performance to 51.31 mAP.
% We show % +9.25 mAP 
% a significant improvement when training WSOD with both clean (annotated in Pascal VOC 07) and noisy, but vetted labels from SBUCaps (51.31 mAP) compared to naively combining clean with noisy labels without vetting (42.06 mAP) or only using clean labels (43.48 mAP). 
Lastly, VEIL's 
% generalizes and 
% its 
gains persist across datasets, object vocabulary, and scale.
%WSOD performance improves with vetting over no vetting as the train dataset is scaled.

To summarize, our contributions are:
\begin{enumerate}[nolistsep,noitemsep]
    \item VEIL, a transformer-based extracted label, visual presence classifier, and
    \item constructing 
    the  \textbf{C}aption \textbf{La}bel \textbf{N}oise dataset.
\end{enumerate}

We find that: \begin{enumerate}[nolistsep,noitemsep]
    \item VEIL outperforms language-conditioned, visual-conditioned, and language-agnostic label noise correction approaches
    in vetting labels from a wide set of in-the-wild datasets for weakly-supervised object detection. 
    \item VEIL enables effective combination of extracted noisy and clean labels.
    \item Even when VEIL is trained on one dataset/category, but applied to another, it shows advantages over baselines. 
\end{enumerate}
\section{Related Work}

\textbf{Vision-language datasets} 
include crowdsourced captions \cite{Young2014Flicker30K, Lin2014MicrosoftCC, huang2016visual,Krishna2016VisualGC} and alt-text written by users to aid visually impaired readers \cite{Sharma2018ConceptualCA,Changpinyo2021Conceptual1P,Radford2021LearningTV,Schuhmann2021LAION400MOD} are widely used for vision-language grounding due to abundance and high visual-text alignment.
There are also large in-the-wild datasets sourced from social media like Reddit \cite{Desai2021RedCapsWI} and user-uploaded captions for photos shared on Flickr \cite{Ordonez2011Im2TextDI}. 
We show the narrative element found in these in-the-wild datasets, captured by the linguistic cues we investigate, impact the ability to successfully train an object detection model.

\textbf{Weakly-supervised object detection} (WSOD) is a multiple-instance learning problem to train a model to localize and classify objects from image-level labels \cite{bilen2016weakly,tang2017multiple,wan2019cmil,gao2019cmidn,ren2020instance,Shao2022DeepLF}.
Cap2Det was the first work to leverage unstructured text accompanying an image for WSOD by predicting pseudo image-level labels from captions \cite{Ye_Zhang_Kovashka_Li_Qin_Berent_2019, Unal2022LearningTO}.
However, Cap2Det cannot operate across novel categories as it 
directly predicts image-level labels and aims to correct  
% The text classifier in Cap2Det was only used when exact string matching produced no labels, which means it mainly corrects 
% Further, Cap2Det targets 
false negatives, %(visually present, not extracted labels)
 not visually absent extracted labels. Detic \cite{Zhou2022DetectingTC} uses weak supervision from 
ImageNet \cite{Deng2009ImageNetAL} and extracts labels from Conceptual Captions (CC) %\cite{Sharma2018ConceptualCA} 
to pretrain an open vocabulary object detection model with a CLIP classifier head. While these approaches succeed in leveraging relatively clean, crowdsourced datasets like COCO, 
Flickr30K and ImageNet, both see lower performance in training with CC \cite{Unal2022LearningTO,Zhou2022DetectingTC}. Other prior work \cite{Gao2021OpenVO} uses  
a pretrained vision-language model 
%(on 14M data) ALBEF \cite{Li2021AlignBF} 
to generate pseudo-bounding box annotations,
% COCO, Visual Genome, and SBUCaps, 
but always requires clean data (COCO), and does not explicitly study the contribution of in-the-wild datasets.

\begin{table*}[t]
\resizebox{\linewidth}{!}{%
    \begin{tabular}{c|c|c|c|c|c|c|c|c|c|c|c|c|c|c|c}
    \hline 
        & \multicolumn{3}{c|}{Label noise} & \multicolumn{2}{c|}{Similar context} & \multicolumn{3}{c|}{Visual defects} & \multicolumn{7}{c}{Linguistic indicators} \\ \hline
        {\scriptsize Dataset} & {\scriptsize \%Vis} & {\scriptsize \%Part} & {\scriptsize \%Abs} & {\scriptsize \%Co-occ} & {\scriptsize \%Sim} & {\scriptsize \%Occl} & {\scriptsize \%Parts} & {\scriptsize \%Atyp} & {\scriptsize \%Beyond} & {\scriptsize \%Past} & {\scriptsize \%Prep} & {\scriptsize \%Non-lit} &  {\scriptsize \%Mod} & {\scriptsize \%Sense} & {\scriptsize \%Named} \\ \hline
        S & 21.5 & 20.0 & 58.5 & 42.5 & 13.2 & 61.6 & 46.3 & 44.6  & 26.0 &  3.0 & 40.5 & 11.0 &  32.0 & 12.0 & 5.0 \\
        R & 29.2 & 12.8 & 57.5 & 15.0 & 4.0 & 21.8 & 22.2 & 49.0 & 19.8 & 3.1 & 5.7 & 9.3 & 26.6 & 18.2 & 10.9 \\
        CC & 32.8 & 16.6 & 50.5 & 30.9 & 12.8 & 36.3 & 24.2 & 57.3 & 27.6 & 2.6 & 31.3 & 5.7 &  25.0 & 8.3 & 2.1 \\
        \hline 

    \end{tabular}
    }
    \caption{Label noise distributions; ``other''/uncommon categories skipped.
    %roughly sum to 100. The rest of the answers are not exclusive.
    %, hence don't sum to 100. 
    Similar context is only annotated %, and the percent is only calculated, 
    for absent objects agreed by both annotators. Visual defects are annotated over examples with full or partial visibility. Linguistic indicators are annotated 
    %and the percent is calculated 
    over examples with visual defects or partial/no visibility. %These percents are averaged over both annotators. 
    Annotation abbreviations, Q1: Label noise as [Vis = Visible, Part = Partially visible, Abs = Absent], Q2: Similar context as [Co-occ = Co-occurring context, Sim = Semantically similar object], Q3: Visual defects as [Occl = Occlusion,  Parts = Key parts missing, Atyp = Atypical], Q4: Linguistic indicators as [Beyond = Beyond the image, Past = Describes the past, Prep = Prepositional phrase, Non-lit = Non-literal use, Mod = Noun modifier, Sense = Different word sense, Named = Named entity]. Datasets abbreviations:  
    [S = SBUCaps, R = RedCaps, CC = Conceptual Captions].}
    \label{tab:stats}
\end{table*}

\textbf{Vision-language pre-training for object detection.} 
Image-text grounding has been leveraged as a pretraining task for open vocabulary object detection \cite{ZSOD_3_Rahman2020ImprovedVA, ZSL_Rec_Diff_ZSOD_Rahman2020ZeroShotOD, zareian2021open, ZSOD_2_OVOD_1_gu2022openvocabulary, regionclip, Du2022LearningTP,wu2023aligning}, followed by bounding box supervision from base classes. 
Some methods distill knowledge from existing pretrained vision-language grounding models like CLIP and ALIGN \cite{Jia2021ScalingUV} to get proposals \cite{Shi2022ProposalCLIPUO} and supervision for object detection \cite{Du2022LearningTP, regionclip}; however, these do not 
% compare clean vs 
study the effect of noisy supervision in a setting without bounding box supervision. 
In contrast, we perform weakly-supervised object detection (WSOD) using noisy image-level labels from captions only. WSOD is a \textbf{distinct task} from open-vocabulary detection and has the \textbf{advantage} of not requiring expensive 
bounding boxes. % on base classes.
We focus on \textbf{rejecting labels} harmful for localization. 

\textbf{Adaptive label noise reduction in classification.}
Adaptive methods reject or correct noisy labels ad-hoc during training. These methods exploit a network's ability to learn representations of clean labels earlier in training. This assumes there are no clear visual patterns in the noisy samples corresponding to a particular corrupted label, leading to their memorization later in training \cite{Zhang2016UnderstandingDL}.
We instead show diverse real-world datasets contain naturally occurring \emph{structured} noise, where in many cases there are visual patterns to the corrupted label.
Large Loss Matters \cite{Kim2022LargeLM} is representative of such adaptive noise reduction methods and we find that it struggles with noisy labels extracted from in-the-wild captions.

\section{Label Noise Analysis and Dataset}%Caption Label Noise (CLaN) Dataset}
\label{sec:analysis}

We analyze what makes large in-the-wild datasets a challenging source of labels for object detection. 

\textbf{Datasets analyzed.} 
\textbf{RedCaps} \cite{Desai2021RedCapsWI} 
consists of 12M Reddit image-text pairs collected from a curated set of subreddits with heavy visual content.
\textbf{SBUCaps} \cite{Ordonez2011Im2TextDI} consists of 1 million Flickr photos with text descriptions written by their owners.
Captions were selected if at least one prepositional phrase and 2 matches with a predefined vocabulary were found.
Conceptual Captions (\textbf{CC}) \cite{Sharma2018ConceptualCA} contains 3M image-alt-text pairs after heavy post-processing: named entities in captions were hypernymized and image-text pairs were accepted if there was an overlap between Google Cloud Vision API class predictions and the caption. 
%This motivates our exploration of VAEL noise.

\textbf{Extracted object labels.} Given a vocabulary of object classes, we extract a label for an image if there is an exact match between the object name and the corresponding caption ignoring punctuation. 
While this strategy will result in some noisy labels, it represents how labels are extracted in prior work \cite{Ye_Zhang_Kovashka_Li_Qin_Berent_2019,Fang2022DataDD} due to the absence of clean annotations. Using gold standard labels (defined next), we calculate the precision of the extracted labels. %Even with CC's heavy post-processing% causes cleaner labels than the other in-the-wild datasets, 
% All of the i
% In-the-wild datasets exhibit very low extracted label precision,
% % compared to COCO. This precision 
% ranging from 0.463 for SBUCaps, 0.596 for RedCaps, to 0.737 for CC, compared to
% % all much lower than 
% 0.948 for COCO (see no vetting precision in Tab. \ref{tab:direct_eval_all}). 
In-the-wild datasets exhibit much lower extracted label precision, with SBUCaps at 0.463, RedCaps at 0.596, and CC at 0.737, in stark contrast to COCO's 0.948 (refer to Tab. \ref{tab:direct_eval_all} for no-vetting precision).

\textbf{Gold standard object labels.} We use \emph{image-level} predictions from a pretrained image recognition model to \emph{estimate} visual presence \textit{gold standard} labels (pseudo-ground-truth) because in-the-wild datasets do not have object annotations. 
We use an object recognition ensemble with the X152-C4 object-attribute model \cite{zhang2021vinvl} 
and Ultralytic  YOLOv5-XL \cite{yolov5}.
This ensemble achieves strong accuracy, 82.2\% on SBUCaps, 85.6\% on RedCaps, and 86.8\% on CC (see Appendix Sec.~\ref{appx:quality_of_pretrained_image_recognition_ensemble}: we annotate a subset to estimate  accuracy).
%Our cross-category experiments show we do not require labels for all classes.
For our analysis of visually absent extracted labels (VAEL), we sample image-caption pairs where the extracted label and gold standard label disagree.
Note we never use bounding-box pseudo labels, only image-level ones. 

\textbf{Caption Label Noise (CLaN) dataset annotations collected.} % \textbf{Noise annotations collected.}
To understand the label noise distribution, we select 100 VAEL examples per dataset (RedCaps, SBUCaps, CC) and annotate four types of information (abbreviations are underlined): 
\begin{itemize}[nolistsep,noitemsep]
    \item (Q1: Label Noise) How much of the VAEL object is present (\underline{vis}ible, \underline{part}ially visible, completely \underline{abs}ent); 
    \item (Q2: Similar Context) If the VAEL object is completely absent, is there traditionally \underline{co-occ}urring context (``boat'' and ``water'') or a semantically \underline{sim}ilar object (e.g. ``cake'' and ``bread'', ``car'' and ``truck'') is present instead; %in the image; 
    \item (Q3: Visual Defects) If the VAEL object is visible/partially visible, is the object  \underline{occl}uded, have key \underline{parts} missing, or have an \underline{atyp}ical appearance (e.g. knitted animal); and
    \item (Q4: Linguistic Indicators) What linguistic cues explain why the VAEL object is mentioned but absent, e.g. the caption discusses events or information \underline{beyond} what the image shows (see %bottom-right ex. in 
    Fig. \ref{fig:concept}), 
    % (``beyond'' in Tab.~\ref{tab:stats})
    describes the \underline{past} (``earlier that day, my \textbf{dog} peed
on a flower'') or the VAEL is: within 
% extracted label is part of 
a \underline{prep}ositional phrase and likely to describe the setting not objects (e.g. ``on a \textbf{train}''),  used in a \underline{non-lit}eral way (``\textbf{elephant} in the room''), a noun \underline{mod}ifying another noun (``car park''), a different word \underline{sense} (e.g. ``bed'' vs ``river bed''), or part of a \underline{named} entity (see 
% top-left example in 
Fig. \ref{fig:concept}). Note multiple linguistic indicators could be used to detect the absent object.
\end{itemize}

Two authors provide the annotations, with Cohen's Kappa agreements of 0.76 for Q1, 0.33 for Q2, 0.45 for Q3, and 0.58 for Q4. We calculate Cohen's Kappa for each option and 
% aggregate agreement through 
compute a weighted average for each question, with weights derived from average option counts
% between the two annotators 
across annotators and the three datasets. We compute the average disagreement as the number of disagreements divided by the number of samples annotated for each question per dataset, averaged over all datasets. The average disagreement is 25.1\% for Q2, 25.3\% for Q3, 14.6\% for Q4. When comparing similar context (``co-occ'' or ``sim'') vs ``no similar context'' 
% merging similar/co-occurring objects (vs none of them) 
for Q2 and any defects (``occl'', ``parts'', ``atyp'') vs ``no defects'' for Q3, disagreement is 28.7\% for Q2, 17.0\% for Q3. The disagreements are fairly low.
% We label the dataset Caption Label Noise, or CLaN.

In Table \ref{tab:stats}, we show what fraction of samples fall into each annotated category, excluding ``Other'', ``Unclear'' and uncommon categories. We average the distribution between the two annotators.

\textbf{Statistics: Label noise.}
We first characterize the visibility of objects flagged as VAELs by the recognition ensemble. SBUCaps has the highest rate of completely absent images (58.5\%), followed closely by RedCaps. 
SBUCaps also has the highest rate of partially visible objects (20\%). 
CC has the highest full visibility (32.8\%),
%followed by RedCaps (29.2\%) and then SBUCaps (21.5\%) where full visibility is 
defined as the object having 75\% or more visibility from a given viewpoint. 
Samples with 
%the high rate of 
absent and partially-visible objects %justifies using pseudo-ground-truth labels from the recognition ensemble; these both 
constitute poor training data for WSOD, and their high rate motivates our VEIL approach. 

\textbf{Statistics: Similar context.}
Certain images with absent objects may be more harmful than others. Prior work shows that models exploit co-occurrences between an object and its context to do recognition, but when this context is absent, performance drops \cite{Singh_Mahajan_Grauman_Lee_Feiszli_Ghadiyaram_2020}. We hypothesize that including images without the actual object %present
and with this contextual bias could hurt localization when supervising detection \textit{implicitly}. Additionally, semantically similar objects may blur decision boundaries.
Different annotators may have different references for similarity or co-occurrence frequency, but our annotators achieve fair agreement ($\kappa=0.33$). In Table \ref{tab:stats}, we find high rates of co-occurring contexts in samples with completely absent VAELs for SBUCaps (42.5\%) and CC (30.9\%).
SBUCaps and CC also have a 12-14\% rate of %semantically 
similar objects present instead of the VAEL. 
% For example, if the image contains co-occurring context (e.g. "bus stop with buildings") while the object (e.g. "bus") is absent, then the inclusion of this example could harm localization.

\textbf{Statistics: Visual defects.}
We hypothesize there may be visual defects that caused the recognition ensemble to miss fully visible objects. Here, we compute the percent of at least \textit{one} visual defect in fully or partially visible samples: 79\% for CC, 
% In CC, 79\% of fully or partially visible objects have a visual defect (not shown in Tab. \ref{tab:stats}),
87\% for SBUCaps, and 69\% for RedCaps. % over the fully or partially visible subset. 
Tab. \ref{tab:stats} has the distribution by visual defect type; this shows that atypical appearance is the most common defect for RedCaps and CC (49\% and 57.3\%). We argue atypical examples constitute poor training data for WSOD, especially when learning from scratch.
The caption context (e.g. ``acrylic illustration of the funny mouse'') may indicate the possibility of a visual defect, further motivating the VEIL design. 

\textbf{Statistics: Linguistic indicators.} Noun modifiers are frequently occurring indicators over all datasets. Prepositional phrases are significant in SBUCaps (40.5\%) and CC (31.3\%).
Need for caption context in vetting is motivated %All datasets contain many VAELs mentioned in contexts going 
by many VAELs being mentioned in contexts going beyond the image, e.g.:
``just got back from the river. friend \textbf{sank his truck pulling his \underline{boat} out}. long story short, rip this beast'' (RedCaps). 
We find prevalent structured noise (pattern to the images associated with a particular noisy label) for indicators like ``noun modifier'' and ``prepositional phrase'' 
due to high levels of occlusion and similar contexts.

\section{Method}
\label{sec:veil}

% MOVED EXTRACTING LABELS TO ANALYSIS SECTION

\begin{figure}[t]
    %\centering
    \includegraphics[scale=0.27]{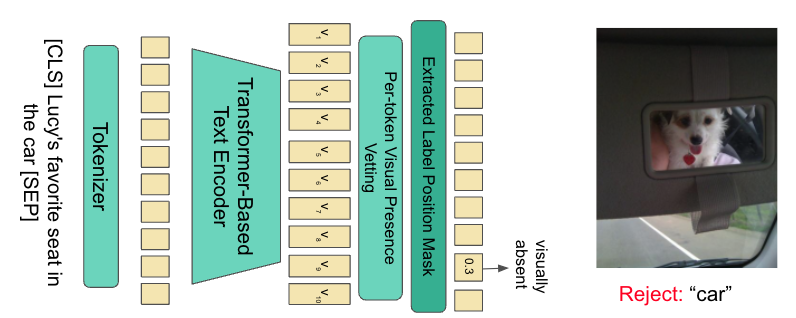}
    \caption{VEIL architecture. 
    In this example, only ``dog'' is an extracted label and it fails the vetting process.
    %After the vetting layer, 
    The masking layer masks visual presence predictions for text tokens not corresponding to an extracted label.
    }
    \label{fig:elavet_arch}
\end{figure}

\textbf{Vetting labels (VEIL).} The extracted label vetting task aims to predict binary visual presence targets (present/absent)
for \textit{each} extracted label in the caption using \textbf{only} the caption context, not the corresponding image. 
We hypothesize there is enough signal in the caption to vet the most harmful label noise. This reduces the model complexity and prevents distractions from the visual modality (similar context). %justification for language only
The method is overviewed in Fig.~\ref{fig:elavet_arch}. Given a caption, WordPiece \cite{wu2016google} produces a sequence of subword tokens $C$; each token is mapped to corresponding embeddings, resulting in $e \in \mathbb{R}^{d\times C}$.
%Our model takes in a sequence of $C$ word token-level caption embeddings. 
These embeddings are passed through a pretrained language model (BERT \cite{BERT}), $h$, which includes multiple layers of multi-head self-attention over tokens in the caption to compute token-level output embeddings $v\in \mathbb{R}^{d\times C}$.
An MLP is applied to these embeddings and the output is a sequence of visual presence predictions per token, $r\in [0,1]^{C}$. 
\begin{gather}
    v = h(e) \\
    r = \sigma(W_2(\tanh(W_1v))
\end{gather}
where $W_1\in \mathbb{R}^{d\times d}$ and $W_2\in \mathbb{R}^{1\times d}$. 

Not all predictions in $r$ correspond to an extracted label, so we use a mask, $M\in [0,1]^{C}$, such that binary cross entropy loss is only applied to predictions/targets associated with the extracted labels. %are used in binary cross entropy loss.  
To train this network, the pseudo-label targets are present, $y_i = 1$, if a pretrained image-level object recognition model also predicts the %same category as the 
extracted label. 
\begin{gather}
    L_{i} = M_i\Big[y_i \log r_i + (1 - y_i) \log(1-r_i)\Big]\\
    L=\frac{1}{M^{T}M}\sum_{i=1}^{C} L_{i}
\end{gather}
While using pretrained object recognition models may appear unfair, %we believe that 
bootstrapping this knowledge to train a language model to predict token-level binary visual presence 
% prediction using only caption input 
has efficiency benefits (no image input required), can generalize to extracted labels outside of the recognition model's vocabulary 
% since visual presence is predicted on a per token level 
(see Sec. \ref{sec:expts} for generalization experiments), and is realistic for WSOD, since detection labels are more limited, whereas many recognition labels exist. 

During \emph{inference}, if an extracted label 
%has been tokenized into 
was mapped to multiple tokens (e.g. ``teddy bear''), the predicted scores are averaged to a single prediction.

%\textbf{Special token.} We test VEIL$_{\text{ST}}$ which inserts a special token {\tt [EM\_LABEL]} before each extracted label in the caption to reduce the model's reliance on category-specific cues and improve generalization to other datasets. We find that it helps only the latter.

\textbf{Weakly-supervised object detection.}
To test the ability of extracted label filtering or correction methods for weakly-supervised object detection, we train MIST \cite{ren2020instance}. MIST extends WSDDN \cite{bilen2016weakly} and OICR \cite{Tang2017MultipleID} which combine class scores for a large number of regions in the image to compute an image-level prediction (used for training).  
%to improve iterative refinement such that multiple instances are not grouped as one.
VEIL uses image-level pseudo-visual presence labels 
% training data 
from the in-the-wild datasets to train the vetting model, and we want to see how its ability to vet labels for WSOD generalizes to unseen data. Thus, we use the test splits of the in-the-wild datasets to train MIST, as they are unseen by all vetting methods. We do not evaluate the WSOD model on these in-the-wild datasets, but on disjoint datasets which have bounding boxes (PASCAL VOC and COCO). 

%\textbf{Weighted sampling.}
% \begin{gather}    
% \end{gather}

%\textbf{Implementation.}
%VEIL is implemented in PyTorch \cite{pytorch} and 
%We use a pretrained BERT encoder \cite{BERT} prior to the per-token visual presence classification layer. 
%We use MIST \cite{ren2020instance} to learn an object detection model using weak supervision from image-level labels extracted from captions. 
%We simulate a batch size of 8.
%for all experiments unless specified otherwise. 
%To address class imbalance during WSOD, 
%caused by in-the-wild datasets' long-tail distributions
%we use the complement of the sub-sampling probability from \cite{Mikolov2013DistributedRO} as weights. 
%We train under different GPU settings due to resource constraints, and use gradient accumulation for some experiments. 
%We used 4 RTX A5000 GPUs and trained for 50k iterations with a batch size of 8, or 100k iterations on 4 Quadro RTX 5000 GPUs with a batch size of 4 and gradient accumulation (parameters updated every two iterations to simulate a batch size of 8).
%We use batch size 8. %See supp for more.
\section{Experiments}
\label{sec:expts}

\begin{table*}[h]
\begin{center}
\small
\begin{tabular}{>{\centering\arraybackslash}m{0.25in}|p{1.65in}|p{0.35in}|p{0.35in}|p{0.35in}|p{0.35in}|p{0.35in}|p{0.35in}|p{0.35in}|p{0.28in}}
\hline
 &  \textbf{Method} & \textbf{S} & \textbf{R} & \textbf{CC} &  \textbf{VIST} & \textbf{VIST-DII} & \textbf{VIST-SIS}& \textbf{COCO} & \textbf{AVG}\\\cline{2-10}

& No Vetting & 0.633 & 0.747 &  0.849 & 0.853 & \underline{0.876} & \underline{0.820} & \textbf{0.973} & 0.822 \\
\hline
&Global CLIP {\scriptsize \cite{Radford2021LearningTV}} &  0.604&	0.583	&0.569	&0.668	&0.625&	0.683&	0.662& 0.628\\
\multirow{-2}{*}{VL}& Global CLIP - E {\scriptsize\cite{Radford2021LearningTV}} &  0.594 &	0.569&	0.534&	0.654&	0.613&	0.660&	0.640 & 0.609\\
 \hline
&   Local CLIP {\scriptsize\cite{Radford2021LearningTV}} &   0.347 & 0.651 & 0.363 & 0.427	&0.476	&0.418	&0.464 &0.449 \\
& Local CLIP - E {\scriptsize\cite{Radford2021LearningTV}} &\underline{0.760} & \underline{0.840} & 0.597 & 0.759	&0.695&	0.812&	0.788 &0.750 \\
\multirow{-3}{*}{ V}& Reject Large Loss  {\scriptsize\cite{Kim2022LargeLM}}
 & 0.667 & 0.790 & 0.831&0.782	&0.794	&0.743&	0.896 &0.786 \\
  \hline
& Accept Descriptive & 0.491 & 0.413 & 0.740 & 0.687&	0.844&	0.264&	0.935 & 0.625\\
%& Accept Narrative & 0.470 & 0.645 & 0.383 & 0.487 &	0.154&	0.757	&0.143 & 0.434 \\
&Reject Noun Mod. & 0.618 & 0.703 & 0.814& 0.823	&0.847	&0.788 &	0.906 & 0.786 \\
%& Reject Noun Mod. (Any) & 0.616 & 0.689 & 0.812 & 0.821	&0.842	&0.782 &	0.900&  0.780\\
 &  Cap2Det {\scriptsize\cite{Ye_Zhang_Kovashka_Li_Qin_Berent_2019}}
 & 0.639 & 0.758 & 0.846& 0.826&	0.854	&0.774	& \underline{0.964}& 0.809\\\cline{2-10}
& VEIL-Same Dataset & \textbf{0.809} & \textbf{0.890} & \textbf{0.909} & \underline{0.871}	&\textbf{0.892}	& 0.816 &	\textbf{0.973} & \textbf{0.884}\\
\multirow{-7}{*}{L}& VEIL-Cross Dataset
 & 0.716 & 0.793 & \underline{0.850} & \textbf{0.875} &	\textbf{0.892}	&\textbf{0.830} &	0.958 & \underline{0.842}\\\hline
\end{tabular}
\end{center}
\caption{Extracted label vetting F1 Performance. %Visual presence ground truth is estimated by an object detection ensemble, X152-C4 \cite{zhang2021vinvl} and YOLOv5-XL \cite{yolov5}, on all datasets except for COCO, where we use existing annotations. 
S=SBUCaps, R=RedCaps. \textbf{Bold} indicates best performance in each column, and \underline{underlined} second-best. (V) signifies method uses the visual modality and (L) uses language.}
%Precision/recall values are shown in our supplementary materials.}
\label{tab:direct_eval_f1}
\end{table*}

We show the ability of VEIL to vet noisy extracted labels, remove structured noise, and outperform language-agnostic filtering and image-based filtering methods. 
We test generalization ability in VEIL through cross-dataset and cross-category experiments. Lastly, we evaluate weakly-supervised object detection settings using only noisy supervision and a combination of noisy and clean supervision.

\subsection{Experiment Details}

We use three in-the-wild image-caption datasets: SBUCaps \cite{Ordonez2011Im2TextDI}, RedCaps \cite{Desai2021RedCapsWI}, Conceptual Captions \cite{Sharma2018ConceptualCA}; and three crowdsourced datasets that fall into descriptive: COCO \cite{Lin2014MicrosoftCC}, VIST-DII \cite{huang2016visual}) and narrative: VIST-SIS \cite{huang2016visual}. 
In-the-wild 
and VIST captions are filtered using substring matching against COCO categories; this creates a subset of image-caption pairs where there is at least one match. This subset is split into 80\%-20\% train-test; see Appendix Sec. \ref{appx:vetting_dataset_details} for image-caption counts. See Sec. \ref{sec:analysis} for details on how pseudo-ground truth visual presence is produced for all datasets except COCO which has object annotations. % datasets where image-level object labels don't exist (e.g. all our datasets except COCO).
The WSOD models are trained on SBUCaps with labels vetted by different methods, and evaluated on PASCAL VOC 2007 test \cite{Everingham2010ThePV} and COCO val 2014 \cite{Lin2014MicrosoftCC}.
%The mAP metric uses $IOU=0.5$. However, when contrasting the image-level classification performance and detection performance we refer to them as Recognition maP and Detection mAP, respectively.

\subsection{Methods Compared}
Since we train and test VEIL on various datasets, we use the convention VEIL-X to signify that VEIL is trained on the \textit{train-split} of X where X is the dataset name. 
%To test the source generalization capacity of VEIL given a data source (SBUCaps), we report VEIL-Cross Dataset results which refers to the best performing VEIL using data sourced from the remaining datasets.
We group the methods we compare against into language-based, visual-based, and visual-language methods. They are category-agnostic, except for Cap2Det \cite{Ye_Zhang_Kovashka_Li_Qin_Berent_2019} and Large Loss Matters (LLM) \cite{Kim2022LargeLM}, both of which must be applied on closed vocabulary. 
\\\textbf{No Vetting} accepts all extracted labels (\emph{recall}=1).
\\\textbf{Global CLIP and CLIP-E} use the ViT-B/32 pretrained CLIP \cite{Radford2021LearningTV} model. To enhance alignment \cite{Hessel2021CLIPScoreAR}, we add the prompt ``A photo depicts'' to the caption and calculate the cosine similarity between the image and text embeddings generated by CLIP. We train a Gaussian Mixture Model with two components on dataset-specific cosine similarity distributions. During inference, we accept image-text pairs with predicted components aligned with higher visual-caption cosine similarity.
For the ensemble variant (CLIP-E), we prepend multiple prompts to the caption
and use maximum cosine similarity.
\\\textbf{Local CLIP and CLIP-E} 
use cosine similarity between the image and the prompt ``this is a photo of a'' followed by the \textbf{extracted label}. This method directly vets the extracted label compared to GlobalCLIP which filters the entire caption. Since the caption context is ignored, this is image-conditioned.
% Only extracted labels are filtered rather than entire captions, making this image-conditioned, not image-language conditioned vetting like Global CLIP. 
Local CLIP-E ensembles prompts.
\\\textbf{Reject Large Loss.} LLM \cite{Kim2022LargeLM} is a language-agnostic adaptive noise rejection and correction method. To test its vetting ability, we simulate five epochs of WSOD training \cite{bilen2016weakly} and consider label targets with a loss exceeding the large loss threshold as ``predicted to be visually absent'' after the first epoch. LLM controls the strength of the rejection rate using the relative delta hyperparameter
% hyperparameter called the relative delta modulates the strength of the rejection rate, and is 
(0.002 in \cite{Kim2022LargeLM}); we use 0.01 and show our ablations in Appendix Sec. \ref{appx:WSOD_Implementation_Details}.
\\\textbf{Accept Descriptive.} We use a descriptiveness classifier \cite{Rai2023ImprovingLO} 
\textit{trained} to predict whether a VIST \cite{huang2016visual} caption comes from the DII (descriptive) or SIS (narrative) split. The input is a multi-label binary vector representing part of speech tags (e.g. proper noun, adjective, verb - past tense, etc) present. % in the caption. 
We accept extracted labels from captions with descriptiveness over 0.5.
\\\textbf{Reject Noun Mod.} Since an extracted label could be modifying another noun (``\underline{car} park''), a simple baseline is to reject 
an extracted label if the POS label is an adjective or is followed by a noun. 
\\\textbf{Cap2Det.} We reject a label if it is not predicted by the Cap2Det \cite{Ye_Zhang_Kovashka_Li_Qin_Berent_2019} classifier.

\begin{table*}[]
\begin{center}
\small 
\resizebox{\linewidth}{!}{%
    \begin{tabular}{c|l|c|c|c|c|c|c|c|c|c|c|c|c|c}
    \hline
        Data & Vetting Method & \multicolumn{2}{c|}{Label noise} & \multicolumn{2}{c|}{Similar context} & \multicolumn{3}{c|}{Visual defects} & \multicolumn{6}{c}{Linguistic indicators} \\ \hline
        &  & {\scriptsize \%Part} & {\scriptsize \%Abs} & {\scriptsize \%Co-occ} & {\scriptsize \%Sim} & {\scriptsize \%Occl} & {\scriptsize \%Parts} & {\scriptsize \%Atyp}  & {\scriptsize \%Mod} & {\scriptsize \%Prep}  & {\scriptsize \%Non-lit} & {\scriptsize \%Sense} & {\scriptsize \%Named} & {\scriptsize \%Beyond} \\ \hline
        %\multicolumn{14}{|c|}{SBUCaps (S)}\\\hline
        \multirow{2}{1.2cm}{SBUCaps} & VEIL-Same Dataset &  \textbf{85.0} & \textbf{94.7} & \textbf{87.0} & \textbf{80.0} & \textbf{81.1} & \textbf{90.6} & \textbf{87.2} & \textbf{95.2} & \textbf{93.9} & \textbf{90.6} & \textbf{100.0} & \textbf{100.0} & \textbf{88.8} \\ 
        % VEIL-Cross Dataset & 45.2 & 72.3 & 67.5 & 63.7 & 55 & 60.4 & 65.6 & 47.8 & 76.6 & 77.5 & 75.6 & 91.6 & 70.8 & 52.8 \\ \hline
        & LocalCLIP-E  & 51.5 & 80.7 & 71.3 & 70.0 & 52.7 & 52.1 & 65.6 & 63.8 & 70.6 & 82.9 & 96.2 & 62.5 & 82.4 \\ \hline
        %\multicolumn{14}{|c|}{RedCaps (R)}\\\hline
        \multirow{2}{1.2cm}{RedCaps} & VEIL-Same Dataset  & \textbf{91.7} & 74.1 & \textbf{71.4} & \textbf{85.7} & \textbf{83.3} & \textbf{89.0} & \textbf{68.3} & \textbf{74.8} & \textbf{90.0} & 66.7 & \textbf{88.9} & 80.9 & 76.3 \\ 
        % VEIL-Cross Dataset & 54.1 & 68.3 & 61.5 & 50 & 45.2 & 58.3 & 74.2 & 50 & 69.2 & 63.3 & 33.3 & 72.2 & 71.4 & 60.4 \\ \hline
        & LocalCLIP-E  & 52.8 & \textbf{78.4} & 40.0 & 38.1 & 47.0 & 45.0 & 23.2 & 68.4 & 63.3 & \textbf{70.8} & 70.6 & \textbf{90.0} & \textbf{76.7} \\ \hline
        %\multicolumn{14}{|c|}{Conceptual Captions (CC)}\\\hline
        \multirow{2}{1.2cm}{CC} & VEIL-Same Dataset  & \textbf{60.6} & 83.0 & \textbf{81.2} & 55.0 & \textbf{54.9} & \textbf{53.6} & \textbf{56.3} & 64.2 & \textbf{73.7} & 81.7 & \textbf{100.0} & - & 77.4 \\ 
        % VEIL-Cross Dataset & 61.6 & 42.8 & 64.1 & 67.4 & 45 & 42.2 & 50 & 65.6 & 76.7 & 54 & 63.3 & 87.5 & - & 45.4 \\ \hline
        & LocalCLIP-E  & 45.0 & \textbf{89.1} & 74.9 & \textbf{57.5} & 49.9 & 50.0 & 24.1 & \textbf{73.3} & 63.9 & \textbf{91.7} & \textbf{100.0} & - & \textbf{86.8} \\ \hline
    \end{tabular}
    }
    \end{center}
    \caption{
    %Vetting 
    VAEL recall on CLaN. Bold indicates best performance per column/dataset. We omit named entity results for CC as it substitutes them with predefined categories (e.g. person, org.).}
    %Top = SBUCaps, middle = RedCaps, bottom = CC.}
    \label{tab:vetting_analysis}
\end{table*}

\subsection{Extracted Label Vetting Evaluation}

\textbf{VEIL selects cleaner labels 
compared to no vetting and other methods, even when evaluated on datasets differing from the training dataset (e.g. trained on Redcaps-Train and evaluated on SBUCaps-Test).}
Tab.~\ref{tab:direct_eval_f1} shows the F1 score which is the harmonic mean of the vetting precision and recall (shown separately in Appendix Sec. \ref{appx:vetting_prec_rec}).  
Most language-based methods, except Accept Descriptive, improve or maintain the F1 score of No Vetting, even though it has perfect recall. Rule-based methods and Cap2Det perform strongly but are outperformed by both VEIL-Same Dataset (trained and tested on the same dataset) and VEIL-Cross Dataset (trained on a different dataset than that shown in the column; we show the best cross-dataset result in this table; see Appendix Sec. \ref{appx:cross_dataset_ablations} for all cross-dataset results). VEIL-Cross Dataset outperforms other language-based approaches, showing VEIL's generalization potential, except on COCO where Cap2Det does slightly better.
Image-and-language-conditioned approaches (Global CLIP/CLIP-E) make label decisions based on the overall caption, so if part of the caption is visually absent, the alignment could be low.
%certain language can affect the alignment even if the object is actually visually present. 
%Table \ref{tab:direct_eval} shows these methods obtain low F1 scores.
Among image-based approaches for label vetting, Local CLIP benefits significantly from using an ensemble of prompts compared to Global CLIP; ensembling prompts improves zero-shot image recognition in prior work \cite{Radford2021LearningTV}. 
Reject Large Loss has the strongest F1 score among the image-based methods, but is worse than VEIL.

\textbf{Using CLaN, we find that VEIL is stronger than CLIP-based vetting at rejecting different forms of label noise.
Captions alone contain cues about noise.}
We hypothesize that LocalCLIP-E would do well at vetting VAELs explained by linguistic cues like ``non-literal'' and ``beyond the image'' as they are likely to have low image-caption cosine similarity.
We also hypothesize that VEIL would do better than LocalCLIP-E at vetting VAELs that are noun modifiers or in prepositional phrases, which can be easily picked up from the caption.
Further, visual noise in the form of similar context but absent/partially visible object (Q2 in CLaN), could be detected by VEIL from linguistic cues like noun modifiers, prepositional phrases, or caption context implying different word sense. However, LocalCLIP-E may be oblivious to the context differing from the VAEL category. 
We evaluate these hypotheses on the CLaN dataset in Tab.~\ref{tab:vetting_analysis}. We omit ``visible'' VAEL samples as these may be pseudo-label errors and the ``past'' linguistic indicator due to too few samples. 
We find VEIL vets truly absent objects for SBUCaps much better than LocalCLIP-E, and comparably for RedCaps or CC. It vets partially visible objects better than LocalCLIP-E by a significant margin; these can be harmful in WSOD which is already prone to part domination \cite{ren2020instance}.
VEIL also recognizes that similar context rather than the actual VAEL category, are present.
VEIL performs better at vetting visible objects that have visual defects
which can be mentioned in caption context 
(``acryllic illustration of \underline{dog}'').
As expected, we find that for all datasets, VEIL vets VAELs from prepositional phrases better than LocalCLIP-E, and noun modifiers for SBUCaps and RedCaps. LocalCLIP-E does better on ``beyond the image'' and non-literal VAELs except on SBUCaps where VEIL excels.

\begin{table}[]
    \centering
    \small
    \begin{tabular}{p{60pt}|p{52pt}|c|c}
    \hline
    Method & Train Dataset & Prec/Rec & F1 \\
    \hline
    No Vetting & - & 0.463 / 1.000 & 0.633 \\ \hline 
    VEIL & SBUCaps & 0.828 / 0.791 & 0.809 \\ \hline 
    VEIL & RedCaps (R) & 0.668 / 0.759 & 0.710 \\
    VEIL & CC & 0.585 / 0.846 & 0.692 \\
    VEIL & R, CC & 0.689 / 0.722 & 0.705 \\ \hline
    %VEIL$_{\text{ST}}$ & R, CC & 0.649 / 0.797 & 0.716 \\ \hline
    % \hline
    % Alignment Model + CLIP Ensemble (max) & RedCaps, WIT \cite{Radford2021LearningTV} & 0.617 / 0.928 & 0.741 \\
    % \hline
    %  Alignment Model + CLIP Ensemble (min) & RedCaps, WIT \cite{Radford2021LearningTV} & 0.814 / 0.676 & 0.738 \\
    % \hline
    LCLIP-E & WIT  & 0.708 / 0.820 & 0.760 \\
    VEIL+LCLIP-E & R,CC,WIT& 0.733 / 0.848 & 0.786 \\ \hline
    \end{tabular}

    \caption{Source generalization of VEIL; vet on SBUCaps. LCLIP-E is LocalCLIP-E. CLIP trained on WIT.} %\cite{Radford2021LearningTV}.}
    % We can improve performance of alignment model by using a special token for visual presence prediction. Furthermore, ensembling with CLIP improves both precision and recall leading to a 2.6 pt increase in the F1 score; the ensemble averages the GMM alignment probability and visual presence probability from the VEIL model. compared to CLIP. Incorporating more datasets improves precision at the cost of recall, however training on multiple datasets with a special token improves recall (*) data used for pretrained CLIP model.}
    \label{tab:cross_dataset_direct_eval}
\end{table}

\textbf{VEIL generalizes across training sources and is complementary to CLIP-based vetting.}
We train VEIL on one dataset (or multiple) and evaluate on an unseen target. We find that combining multiple sources improves precision (Tab.~\ref{tab:cross_dataset_direct_eval}). 
%To better utilize caption context, we test VEIL$_{\text{ST}}$ which predicts visual presence using a special token {\tt [EM\_LABEL]}. We find that this improves F1 performance. Lastly, 
We also try ensembling by averaging predictions between LocalCLIP-E and VEIL-Cross Dataset and find that both are complementary; that is, the ensemble has better precision and recall compared to VEIL-Cross Dataset or LocalCLIP-E alone. 
There is still a significant gap between VEIL-Same Dataset and even the ensembled model in terms of precision and F1. We leave improving source generalizability to future research.

\begin{table}[]
    \centering
    \small
    \begin{tabular}{c|c|c}\hline
Method  & Prec/Rec & F1 \\\hline
No Vetting & 0.323 / 1.000 & 0.488 \\\hline
ID & 0.651 / 0.656  & 0.654 \\\hline
OOD & 0.585 / 0.556 & 0.570 \\\hline
    \end{tabular}
    \caption{VEIL category generalization on SBUCaps.}
    %-ID. }
    \label{tab:cross_category}
\end{table}

\textbf{VEIL produces cleaner labels even on unseen object categories.}
We define an in-domain category set (ID) of 20 randomly picked categories from COCO \cite{Lin2014MicrosoftCC}, and an out-of-domain category set (OOD) consisting of the 60 remaining categories. We restrict the labels using these limited category sets and create two train subsets, ID and OOD from SBUCaps \textit{train} and one ID test subset from SBUCaps \textit{test}. 
We find that transferring VEIL-OOD to unseen categories improves F1 score compared to no vetting as shown in Table \ref{tab:cross_category}. Additionally, VEIL-OOD has higher precision (0.59) compared to LocalCLIP-E (0.53) which was trained on millions of image-captions. This indicates an ability to reject false positive labels from unseen classes.
We hypothesize training on more categories could improve category generalization, but leave further experiments to future research.

\textbf{Why can VEIL generalize? }%Generalization.} These two generalization experiments raise questions on how VEIL can generalize across training sources or out-of-distribution scenarios. 
 We hypothesize that linguistic indicators explaining the visually absent label can be found in captions across datasets and \textit{can} be independent of the object category: past tense, prepositional phrase, noun modifier, and named entities are all represented within BERT \cite{BERT}, which we finetune in VEIL. To evaluate the effect of linguistic indicators in generalization, we compute the \emph{distance} between the linguistic indicator distributions for each dataset pair in CLaN. We compute the correlation between the \emph{distance} and cross-dataset performance. %(all results shown in Appendix Sec. \ref{appx:cross_dataset_ablations})
 We observe a moderately strong negative Pearson correlation ($\rho = -0.62$). % between linguistic indicator distribution distance and cross-dataset performance.
% ; this means that similar (low distance) linguistic indicator distributions correlate with higher cross-dataset performance. 
This indicates that VEIL implicitly learns associations between linguistic indicators 
 and VAELs which can help in generalizing.
%, without being instructed to learn them.

% \begin{table}[t] %[h]
% \begin{center}
% \small
% \begin{tabular}{p{1.72in}|p{0.75in}|p{0.75in}}
% \hline
% \textbf{Method (Train Data Size in Thousands)} &\textbf{VOC Det. mAP} ($\Delta$) &\textbf{VOC Rec. mP} ($\Delta$) \\
% \hline
% No Vetting & 31.2 & 65.3 \\
% GT* & 40.0 \textcolor{green}{(28.2\%)} & 69.0 \textcolor{green}{(5.7\%)}\\
% Large Loss \cite{Kim2022LargeLM} &  \textcolor{red}{()} &   \textcolor{red}{(\%)}\\
% LocalCLIP-E \cite{Radford2021LearningTV} & 37.1 \textcolor{green}{(18.9\%)} & 70.7 \textcolor{green}{(8.2\%)}\\
% VEIL$_{\text{ST}}$-R,CC & \underline{37.8} \textcolor{green}{(21.2)} & 71.4 \textcolor{green}{(9.2\%)} \\
% VEIL-SBUCaps & \textbf{40.5} (29.8\%) & 74.3 (13.8\%) \\
% \hline
% \end{tabular}
% \end{center}
% \caption{Impact of vetting on WSOD performance on VOC-07 and COCO-14 datasets. There is a significant difference in detection and recognition on VOC-07 illustrated by $\Delta$, relative performance change w.r.t. VEIL-SBUCaps on the same column. This highlights that VEIL variants filter out labels harmful to localization. (GT*) directly vets labels using the pretrained object detectors which were used to train VEIL.} 
% \label{tab:det_v_recognition}
% \end{table}
\begin{table}[t] %[h]
\begin{center}
\small
\begin{tabular}{p{1.72in}|p{0.22in}|p{0.2in}|p{0.2in}}
\hline
\textbf{Method} & \scriptsize{ VOC Det. $\text{mAP}_{50}$}  &\scriptsize{VOC Rec. mAP} &\scriptsize{COCO Det $\text{mAP}_{50}$}\\
\hline
GT* (upper bound) & 40.0 & 69.0 & 9.2\\\hline
No Vetting & 31.2 & 65.3 & 7.7\\
Large Loss \cite{Kim2022LargeLM} & 30.9 & 65.3 & 7.5\\
LocalCLIP-E \cite{Radford2021LearningTV} & 37.1  & 70.7 & 7.9\\
%VEIL$_{\text{ST}}$-R,CC & \underline{37.8}  &\underline{71.4} & \underline{8.6}\\
VEIL-R,CC & \underline{37.8}  &\underline{71.4} & \underline{8.6}\\
VEIL-SBUCaps & \textbf{40.5}  & \textbf{74.3} & \textbf{10.4}\\
\hline
\end{tabular}
\end{center}
\caption{Impact of vetting on WSOD performance on VOC-07 and COCO-14. 
%There is a significant difference in detection and recognition on VOC-07 illustrated by $\Delta$, relative performance change w.r.t. no vetting on the same column. 
%VEIL variants filter out labels harmful to localization. 
(GT*) directly vets labels using the pretrained recognition models used to train VEIL.}
%Bold indicates best performance in column excluding GT*.} 
\label{tab:det_v_recognition}
\end{table}

\subsection{Impact on Weakly-Sup. Object Detection}

%\textbf{Comparison between High Performing Methods in Extracted Label Vetting}
We select the most promising vetting methods from the previous section and use them to vet labels from an in-the-wild dataset's, SBUCaps, unseen (\textit{test}) split and then train WSOD models using the vetted labels. Then, these WSOD models are evaluated on detection benchmarks like VOC-07 and COCO-14.
We evaluate two different VEIL methods, VEIL-SBUCaps and VEIL% $_{\text{ST}}$
-RedCaps,CC  
% Both vet labels from SBUCaps and use them to train WSOD, but the vetting method is trained on either (1) SBUCaps or (2) RedCaps and CC captions.
to demonstrate the generalizability of VEIL on WSOD.
Note that we relax Large Loss Matters \cite{Kim2022LargeLM} to \textit{correct} visually absent extracted labels, in addition to unmentioned but present objects (false negatives).
After vetting, we remove any images without labels and since category distribution follows a long-tail distribution, we apply weighted sampling \cite{Mikolov2013DistributedRO}.
We train MIST \cite{ren2020instance} for 50K iter. with batch size 8. %for each method. 

\textbf{VEIL vetting leads to better detection and recognition capabilities than vetting through CLIP, or an adaptive label noise correction method (Large Loss Matters).} We find that VEIL-SBUCaps performs the best as shown in Tab.~\ref{tab:det_v_recognition}.
In particular, it boosts the detection performance of No Vetting by 9.3\% absolute and 29.8\% relative gain (40.5/31.2\% mAP) on VOC-07 and by 35\% relative gain (10.4/7.7\% mAP) on COCO.
Interestingly, VEIL-SBUCaps and VEIL-Redcaps,CC have a similar performance improvement, despite VEIL-Redcaps,CC (best VEIL cross-dataset result on SBUCaps) having poorer performance than Local CLIP-E in Tab.~\ref{tab:cross_dataset_direct_eval}. 

\textbf{VEIL generalizes from its bootstrapped data}. Directly using predictions from the pretrained object recognition model (used to produce visual presence targets for VEIL at the image level) to vet (GT* method in Tab.~\ref{tab:det_v_recognition}) 
performs worse than VEIL % in both detection and recognition \
as shown by 40.5 mAP vs 40.0 mAP on VOC 
% for detection and 74.3 vs 69.0 on VOC for recognition. This also extends to COCO, where we observe 
and 10.4 mAP vs 9.2 mAP on COCO. 
% VEIL generalizes from its bootstrapped data: w
We speculate that learning to identify label noise is an easier task than categorizing different objects; furthermore, image recognition models could still select samples that might be harmful for learning localization (similar contexts, occlusion, etc). % so this could generalize better to unseen data
% as it is learning essential features akin to knowledge distillation from a teacher model 
 The image recognition model may also wrongly reject clean labels. %These differences in predictions would be due to the difference in input (caption rather than image), where possibly, VEIL needs less data to understand visual presence because captions will have less variation than natural images.  
 We leave further exploration to future research.

\textbf{Structured noise negatively impacts localization.} Using the CLaN dataset, we observe
%a number of examples with structured noise. 
one type of structured noise found from extracting labels from prepositional phrases, specifically where images were taken inside vehicles. We hypothesize such structured noise would have significant impact on localization for the vehicle objects. We use CorLoc to estimate the localization ability on vehicles in VOC-07 (``aeroplane'', ``bicycle'', ``boat'', ``car'', ``bus'', ``motorbike'', ``train''). We observe a CorLoc of 60.2\% and 54.1\% for VEIL-SBUCaps and LocalCLIP-E, respectively. 
%For another supercategory, animal, we observe a smaller improvement (57.9 vs 56.8) from LocalCLIP-E and VEIL-SBUCaps. 
This shows structured noise can have a strong impact on localization.

\begin{table}[]
    \centering
    \small
    \begin{tabular}{c|c|c|c|c}\hline
        Clean Labels & Noisy Labels & WS & Vetting & $\text{mAP}_{50}$\\\hline
         % & \checkmark &  &  & 16.67 \\\hline
        \checkmark &  &  & n/a & 43.48 \\\hline
        $\checkmark$ & \checkmark &  &  & 42.06 \\\hline
        $\checkmark$ & $\checkmark$ &  & $\checkmark$ & 51.31 \\\hline
        $\checkmark$ & $\checkmark$ & $\checkmark$  &$ \checkmark$ & 54.76 \\\hline

    \end{tabular}
    \caption{Mixed supervision from clean (VOC-07 trainval) and noisy labels (SBUCaps).
    Eval on VOC-07 test.}
    %WS stands for weighted sampling.} %Note vetting can only be applied on labels extracted from captions, hence are applied only to the noisy labels column.}
    \label{tab:mixed_supervision}
\end{table}

\textbf{Naively mixing clean and noisy samples without vetting for WSOD leads to worse performance than only using clean samples. Vetting
% and weighted sampling 
in-the-wild samples (noisy) with VEIL is essential to improving performance.} We study how vetting impacts a setting where labels are drawn from both annotated image-level labels from 5K VOC-07 train-val \cite{Everingham2010ThePV} (clean) and 50K in-the-wild SBUCaps \cite{Ordonez2011Im2TextDI} captions (noisy). In Tab.~\ref{tab:mixed_supervision} we observe that naively adding noisy supervision to clean supervision actually hurts performance  %(-3.2\%) 
compared to only using clean supervision. After vetting the labels extracted from SBUCaps  \cite{Ordonez2011Im2TextDI} using VEIL-SBUCaps, we observe that the model sees a 17.9\% relative improvement (51.31/43.48\% mAP) compared to using only clean supervision from VOC-07. We see further improvements when applying weighted sampling (WS) to the added, class-imbalanced data (54.76/51.31\% mAP).

\textbf{VEIL improves WSOD performance even at scale.} 
We sampled the held-out RedCaps dataset in increments of 50K samples up to a total of 200K samples. For each scale, we train two WSOD models with weighted sampling using the unfiltered samples and those vetted with VEIL-SBUCaps,CC. 
The mAP at 50K, 100K, 150K, and 200K samples is 4.2, 10.7, 12.0, 12.9 with vetting and 1.9, 8.2, 10.6, 10.4 without vetting.
The non-vetted model's performance declines after 150K samples. 
% Even when VEIL is trained on other datasets, this
% This indicates vetting can adapt to scale better even when VEIL is trained on other datasets. The 
This trend suggests that vetting will continue outperforming no-vetting when dataset sizes increase.

\section{Conclusion}
We released the Caption Label Noise (CLaN) dataset where we annotated types of visually absent extracted labels 
and linguistic indicators of noise in 300 image-caption pairs from three in-the-wild datasets.
Using CLaN, we find that caption context can be used to vet (filter) extracted labels from caption context. We proposed VEIL, a lightweight text model which is trained to predict visual presence using pseudo labels sourced from two pretrained models for recognition. VEIL outperformed nine baselines representative of current noise filtering techniques that could be adapted for captions.

% Our simple technique for filtering (VEIL) avoids some of the biases and limitations of prior methods that could be adapted for use in filtering, e.g. CLIP and Large Loss. 
We demonstrate three key findings specific to vetting for WSOD: (1) there is a distinct advantage in learning to filter 
as opposed to filtering using pseudo-ground truth visual presence labels; 
% that were used to train VEIL
(2) vetting noisy labels is necessary to improve performance  
% not only when training exclusively on noisy data but also
when combined with a clean data source (existing image recognition and detection datasets); (3) structured noise such as noun modifiers and prepositional phrases (e.g. ``car window'', ``on a boat'') has a disproportionate impact on localization and was difficult to detect using visual-based methods like CLIP and Large Loss Matters.
% through the higher performance of VEIL-vetted WSOD compared to CLIP-vetted WSOD. 
This last finding 
implies that %(1) popular methods to filter image-caption pairs like CLIP can miss certain types of noise and (2)
not all noise is equal in impact. CLaN is a starting point for this type of analysis and further research is needed to expand noise categories and measure the impact of the different types of noise.% and measure their impact. 

% Considering that there are existing clean human-annotated datasets, we show that adding extracted labels from in-the-wild captions (noisy) is ineffective without vetting. We leave exploring further analysis on why this happens to future research.

% for many vision tasks. 
% Our work can enhance value in (1) the categorization of label noise especially because those categories can be used as guidance for a VLM, and in (2) establishing diverse baselines for the vetting task.

% Old conclusion
% We showed visually absent extracted labels are common in the wild, VEIL which uses language context to infer if mentioned objects are visually present, and the benefits of its vetting.
%generalizes across datasets and categories, and its benefits persist when adding noisy to clean data. 
\section*{Limitations}

We identify the following limitations of our work. 
First, we assume that captions from SBUCaps, RedCaps, CC cover most in-the-wild caption types.
Second, while VEIL shows promise in generalizing across datasets, there is a performance drop due to label noise distribution differences between datasets. For example, Table 1 shows differences in linguistic indicator distributions across datasets. Since VEIL relies on caption context, it will be sensitive to such changes as shown by our generalization analysis in Sec. \ref{sec:expts}.
Third, VEIL also shows that it can filter unseen object categories (Table 5), however, its performance is noticeably below VEIL-ID which was trained on those object categories. This would be an interesting future direction for research.
Fourth, we noticed that MIST (WSOD method) was highly sensitive to learning rate and that Large Loss Matters was highly sensitive to hyperparameters. We have included these results in A.4.
Fifth, VEIL is sensitive to the gold labels used for training. % VEIL. 
We found that using weaker models (VinVL) to produce labels for VEIL will lead to suboptimal vetting and WSOD results compared to using a stronger model (YOLOv5). 

Lastly, generative vision-language models such as GPT4-V \cite{Achiam2023GPT4TR} open an opportunity to reject noisy labels as well. We think our work would be useful in aiding GPT4-V; a prompt defining noisy samples could use criteria from CLaN (e.g. types of object visibility, visual defects, and linguistic indicators categories). We believe VEIL still serves as a \textbf{lightweight} method to vet labels and could be trained using pseudo-visual presence labels from any source, including generative vision-language models.

% Acknowledgement should be after limitations? Most NLP papers didn't have an acknowledgement but this one did and it came after the limitations: https://aclanthology.org/2023.acl-long.188.pdf

\noindent
\textbf{Acknowledgement.} This work was supported by National Science Foundation Grants No. 2006885 and 2046853, and University of Pittsburgh Momentum Funds.

% \bibliographystyle{ieee_fullname}
\bibliography{anthology, aaai24, egbib}

 \appendix

 \section{Appendix}
\label{sec:appendix}
%We provide supplemental materials to our main text. 

In Section \ref{appx:quality_of_pretrained_image_recognition_ensemble}, we evaluate the quality of the pretrained image recognition ensemble and in Section \ref{appx:vetting_dataset_details}, we present additional dataset details such as counts. In Section \ref{sec:expts} from the main text, we provided only vetting F1 scores in Table \ref{tab:direct_eval_f1} over multiple methods, so in Section \ref{appx:vetting_prec_rec} we provide a detailed table of the vetting precision and recall for the same methods. Furthermore in Section \ref{appx:cross_dataset_ablations}, we show more comprehensive cross-dataset ablations, such as adding more training datasets and training with a special token. 

We discuss our hyperparameter selection for WSOD in further detail in Section \ref{appx:WSOD_Implementation_Details} and show additional metrics of the WSOD models on the COCO-14 benchmark presented in the main text in Section \ref{appx:wsod_benchmarking_on_additional_coco_metrics}.

Finally in Section \ref{appx:qual_examples}, we showcase the vetting ability of VEIL in comparison to other approaches through qualitative results, along with additional examples from the WSOD models trained using vetted training data.

\subsection{Quality of Pretrained Image Recognition Ensemble}
\label{appx:quality_of_pretrained_image_recognition_ensemble}

\begin{table}[!ht]
    \centering
    \begin{tabular}{|c|c|c|}
    \hline
        \textbf{Method} & \textbf{Precision} & \textbf{Recall} \\ \hline
        VinVL Detector & 0.725 & 0.356 \\ \hline
        YOLOv5 & 0.874 & 0.910 \\ \hline
        Ensemble & 0.803 & 0.917 \\ \hline
    \end{tabular}
    \caption{Precision and recall of image recognition models on COCO-14 \cite{Lin2014MicrosoftCC}.}
    \label{tab:coco_ensemble}
\end{table}

\begin{table}[!ht]
    \centering
    \begin{tabular}{|c|c|c|c|} 
    \hline
        \textbf{Method} & \textbf{RedCaps} & \textbf{CC} & \textbf{SBUCaps}\\ \hline
        VinVL detector & 0.764 & 0.572 & 0.688\\ \hline
        YOLOv5 & 0.848 & 0.848 & 0.824 \\ \hline
        Ensemble & 0.856 &  0.868 &  0.822\\  \hline
    \end{tabular}
    \caption{Visual presence accuracy of in-the-wild datasets using annotated examples as ground truth.}
    \label{in_the_wild_ens}
\end{table}

Since we used vision-language datasets without any object annotations, we have no way of knowing whether an object mentioned in the caption is present in the image. To keep our method scalable and datasets large, we used object predictions from pretrained image recognition models to produce visual presence pseudo labels for extracted labels. We test the VinVL detector \cite{zhang2021vinvl} and YOLOv5 detector \cite{yolov5}, and their ensemble (aggregating predictions) on COCO-14 Image Recognition in Table \ref{tab:coco_ensemble} and a visual presence annotated subset in Table \ref{in_the_wild_ens}. For the latter, per dataset we annotated the visual presence of 50 extracted labels from unique images for each category. We used the following randomly selected VOC \cite{Everingham2010ThePV} categories: elephant, truck, cake, bus, and cow. We found that while the ensemble variant and the VinVL detector are worse than YOLOv5 in image recognition on a common benchmark, COCO-14, the ensemble performs better than the single models on visual presence. Since this is the task we aim to do, we select the ensemble model to generate visual presence targets. Additionally, these results indicate there is still significant noise in using these models to generate pseudo labels, so using these pretrained image recognition models is not the same quality as human annotations. Despite this, VEIL still successfully harnesses these noisy targets to reason about visual presence from captions.

\subsection{Vetting Dataset Details}
\label{appx:vetting_dataset_details}
\begin{table}[h]
    \small
    \centering
    \begin{tabular}{c|c|c} \hline
         \textbf{Dataset} & \textbf{Train } & \textbf{Test} \\ \hline
        VIST & 20339 & 5086 \\ \hline
        VIST-DII & 12106 & 3028 \\ \hline
        VIST-SIS & 8233 & 2060 \\ \hline
        COCO & 216096 & 94004 \\\hline
        SBUCaps & 166986 & 41747 \\\hline
        RedCaps & 845333 & 211334 \\\hline
        CC      &  350043 & 87511 \\\hline
    \end{tabular}
    \caption{The number of samples per split and dataset after filtering captions based on exact match with COCO objects. Note VIST and COCO have multiple captions per image; for the sake of vetting, we evaluate on extracted labels from all captions.}
    \label{tab:datacount}
\end{table}

While the overall image-text pairs are 12M pairs for RedCaps, 3M pairs for CC, 1M for SBUCaps, 500K pairs for COCO, 40K and 60K pairs for VIST-DII and VIST-SIS, respectively, after extracting labels using exact match with COCO categories, there are a number of captions which don't have any matches. We filter out those captions. In Table \ref{tab:datacount} we provide counts after filtering for both vetting train and test splits of each dataset.

\begin{table}[t]
    \centering
    \begin{tabular}{c|c}
        Relative Delta & Pascal VOC-07 $\text{mAP}_{50}$  \\\hline
         0.002 & 28.25 \\
         0.01 & 30.93\\
         0.05 & 28.11 \\
    \end{tabular}
    \caption{Relative delta hyperparameter ablation}
    \label{tab:rel_delta}
\end{table}

\begin{figure}[t]
  \centering
  %\subfigure[This figure shows the impact of pseudo-labeled data size for VEIL-SBUCaps on its performance on SBUCaps-Test VPEL Detection. VEIL only needs 10\% of the data to beat No Vetting,
    %(XX vs 0.633 F1),     and 50\% of the labeled data (80K samples) to beat CLIP.]{
    \includegraphics[width=0.47\textwidth]{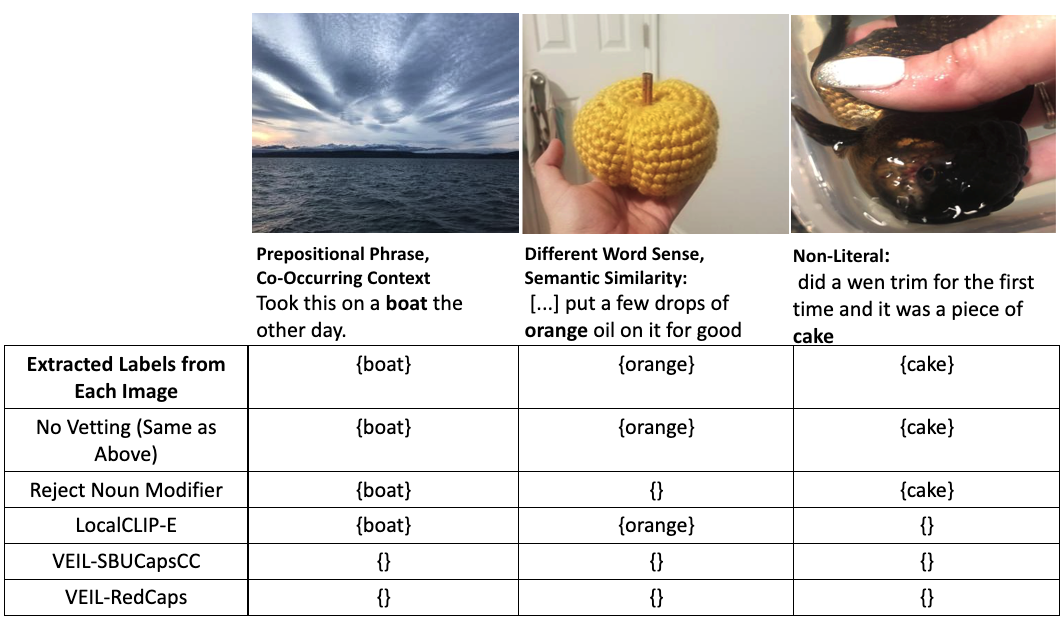}
    % %} \subfigure[Caption for subfigure 2]{
    % \includegraphics[width=0.2\textwidth]{iccv2023AuthorKit/figures/images/.png}
    % %}
    % \includegraphics[width=0.2\textwidth]{iccv2023AuthorKit/figures/images/.png}
  \caption{Qualitative examples of extracted labels after vetting on RedCaps-Test. These are additional completely absent VAEL examples from CLaN with their linguistic indicators and similar context annotations, and only VEIL-based methods are able to overcome these three noise types.}
\label{fig:vetting_quals}
\end{figure}

\subsection{Vetting Precision/Recall}
\label{appx:vetting_prec_rec}

Table \ref{tab:direct_eval_f1} in the main text showed the F1 on the extracted label vetting task, from twelve methods. In Table \ref{tab:direct_eval_all} here, we separately show Precision and Recall on the same task.

\begin{table*}[h]
\begin{center}
\small 
\begin{tabular}{>{\centering\arraybackslash}c|l|c|c|c|c|c|c}
\hline
\multicolumn{1}{c}{} & \multicolumn{1}{|c}{} & \multicolumn{2}{|c}{\textbf{SBUCaps}} & \multicolumn{2}{|c}{\textbf{RedCaps}} & \multicolumn{2}{|c}{\textbf{Conceptual Captions}} \\\cline{2-8}
 & \textbf{Method} &  \textbf{PREC / REC} &  \textbf{F1} &  \textbf{PREC / REC} &  \textbf{F1} &  \textbf{PREC / REC} &  \textbf{F1} \\
\cline{2-8}
& No Vetting & 0.463 / 1.000 & 0.633 & 0.596 / 1.000 &  0.747 & 0.737 / 1.000  & \underline{0.849} \\
\hline
&Global CLIP %\cite{Radford2021LearningTV} 
& 0.531 / 0.700 & 0.604 & 0.618 / 0.551 & 0.583 & 0.753 / 0.458 & 0.569 \\
\multirow{-2}{*}{\scriptsize{VL}}& Global CLIP - E %\cite{Radford2021LearningTV} 
& 0.526 / 0.683 & 0.594 & 0.625 / 0.522 & 0.569 & 0.745 / 0.417 & 0.534 \\
 \hline
&   Local CLIP %\cite{Radford2021LearningTV} 
& 0.588 / 0.246 &  0.347 & 0.723 / 0.591 & 0.651 & 0.750 / 0.240 & 0.363 \\
 & Local CLIP - E %\cite{Radford2021LearningTV} 
& \underline{0.708} / 0.820 & \underline{0.760} & \underline{0.770} / 0.924 & \underline{0.840} & \underline{0.842} / 0.462 & 0.597 \\
\multirow{-3}{*}{\scriptsize{V}}& Reject Large Loss % \cite{Kim2022LargeLM}
 & 0.530 / \textbf{0.898} & 0.667 & 0.700 / 0.908 & 0.790 & 0.806 / 0.858 & 0.831 \\
  \hline
& Accept Descriptive & 0.449 / 0.542 & 0.491 & 0.561 / 0.326 & 0.413 & 0.739 / 0.741 & 0.740 \\
&Reject Noun Mod. & 0.517 / 0.769 & 0.618 & 0.644 / 0.776 & 0.703 & 0.765 / 0.870 & 0.814 \\
 &  Cap2Det %\cite{ye_2019_cap2det}
 & 0.500 / \underline{0.884} & 0.639 & 0.633 / \textbf{0.945} & 0.758 & 0.758 / \textbf{0.956} & 0.846 \\\cline{2-8}
& VEIL-Same Dataset & \textbf{0.828} / 0.791 & \textbf{0.809} & \textbf{0.855} / \underline{0.929} & \textbf{0.890} & \textbf{0.884} / \underline{0.935}& \textbf{0.909}\\

\multirow{-7}{*}{\scriptsize{L}}& VEIL-Cross Dataset
 & 0.636 / 0.811 & 0.713 & 0.747 / 0.847 & 0.793 & 0.834 / 0.866 &	\underline{0.850} \\
\hline\hline
\multicolumn{1}{c}{\textbf{}} & \multicolumn{1}{|c}{\textbf{}} & \multicolumn{2}{|c}{\textbf{VIST}} & \multicolumn{2}{|c}{\textbf{VIST-DII}} & \multicolumn{2}{|c}{\textbf{VIST-SIS}} \\
\cline{2-8}
 & \textbf{Method} &  \textbf{PREC / REC} &  \textbf{F1}  &  \textbf{PREC / REC} &  \textbf{F1}  &  \textbf{PREC / REC} &  \textbf{F1} \\
\cline{2-8}
 & No Vetting & 0.744 / 1.000 & 0.853 & 0.779 / 1.000 & 0.876 & 0.695 / 1.000 & \underline{0.820}\\ \hline
& Global CLIP %\cite{Radford2021LearningTV} 
& 0.772 / 0.589 & 0.668  & 0.788 / 0.518 & 0.625 & 0.754 / 0.624 & 0.683 \\
\multirow{-2}{*}{\scriptsize{VL}} & Global CLIP - E %\cite{Radford2021LearningTV} 
& 0.769 / 0.569 & 0.654 & 0.785 / 0.504 &  0.613 & 0.741 / 0.595 & 0.660 \\\hline
& Local CLIP %\cite{Radford2021LearningTV} 
& 0.752 / 0.298 & 0.427 & 0.787 / 0.341 & 0.476 & 0.738 / 0.292 & 0.418 \\
& Local CLIP - E %\cite{Radford2021LearningTV}  
& \textbf{0.874} / 0.671 &  0.759  & \textbf{0.886} / 0.572 & 0.695 & \textbf{0.833 }/ 0.793 & 0.812\\ 
\multirow{-3}{*}{ \scriptsize{V}}& Reject Large Loss  %\cite{Kim2022LargeLM} 
& 0.755 / 0.811 &	0.782 &	0.792 /	0.796 &	0.794 &	0.700 / 0.791&	0.743 \\ \hline
& Accept Descriptive  & 0.755 / 0.631 & 0.687 & 0.784 / 0.913 & \underline{0.844 }& 0.686 / 0.163 & 0.264\\
& Reject Noun Mod. & 0.775 / 0.879 & 0.823 & 0.813 / 0.883 & 0.847 & 0.716 / 0.875 & 0.788 \\
& Cap2Det %\cite{ye_2019_cap2det} 
& 0.781 / 0.877 & 0.826 & 0.823 / 0.887 & 0.854 &  0.704 / 0.859 &0.774\\ \cline{2-8}
& VEIL-Same Dataset & 0.789 / \textbf{0.971} & \underline{0.871} & 0.819 / \textbf{0.992} & \textbf{0.892 }& 0.690 / \textbf{0.998} & 0.816 \\
\multirow{-7}{*}{\scriptsize{L}}& VEIL-Cross Dataset & \underline{0.835 }/ \underline{0.920} & \textbf{0.875} & \underline{0.870} / \underline{0.915}
& \textbf{0.892} & \underline{0.765 }/\underline{ 0.920} & \textbf{0.830} \\\hline \hline 
\multicolumn{1}{c}{\textbf{}} & \multicolumn{1}{|c|}{\textbf{}} & \multicolumn{2}{c|}{\textbf{COCO}} & \multicolumn{4}{c}{} \\
\cline{2-4}
 & \textbf{Method} &  \textbf{PREC / REC} &  \textbf{F1} \\\cline{2-4}
& No Vetting & 0.948 / 1.000 & \textbf{0.973 }\\ \cline{1-4}
& Global CLIP %\cite{Radford2021LearningTV} 
& 0.945 / 0.509 & 0.662  \\
\multirow{-2}{*}{\scriptsize{VL}} & Global CLIP - E %\cite{Radford2021LearningTV}  
& 0.931 / 0.487 &  0.640 \\\cline{1-4}
& Local CLIP% \cite{Radford2021LearningTV} 
& 0.951 / 0.307 & 0.464  \\
& Local CLIP - E % \cite{Radford2021LearningTV} 
& 0.972 / 0.663 & 0.788 \\ 
\multirow{-3}{*}{\scriptsize{V} } 
 & Reject Large Loss  %\cite{Kim2022LargeLM} 
 & 0.963 / 0.837	& 0.896 \\ \cline{1-4}
& Accept Descriptive  &  0.948 / 0.923 & 0.935 \\
& Accept Narrative  & 0.942 / 0.077 & 0.143 \\
& Reject Noun Mod. & 0.958 / 0.859 & 0.906 \\
& Cap2Det %\cite{ye_2019_cap2det} 
&\textbf{ 0.978} / \underline{0.950} & \underline{0.964} \\ \cline{2-4}
& VEIL-Same Dataset & 0.948 / \textbf{1.000} & \textbf{0.973} \\
\multirow{-7}{*}{\scriptsize{L}}& VEIL-Cross Dataset & \underline{0.975} / 0.942 & 0.958 \\\cline{1-4}
\end{tabular}
\end{center}
\caption{Extracted label vetting evaluation metrics. Bold indicates best result in column, and in the recall columns No Vetting is excluded as it always has perfect recall.}
\label{tab:direct_eval_all}
\end{table*}

\subsection{Cross-Dataset Ablations}
\label{appx:cross_dataset_ablations}

\begin{table*}[!ht]
    % \centering
% \small
    \begin{tabular}{c|c|c|c|c|c}
    \hline
        \textbf{Train Dataset(s)} & ST & \textbf{DII-VIST } & \textbf{SIS-VIST} & \textbf{COCO } & \textbf{VIST } \\ \hline
        No Vetting  && 0.779 / 1.000 & 0.695 / 1.000 & 0.948 / 1.000 & 0.741 / 1.000 \\ \hline
        SBUCaps   && 0.895 / 0.717 & 0.831 / 0.609 & 0.979 / 0.647 & 0.878 / 0.690  \\ \hline
        RedCaps (R) &  & 0.865 / 0.794 & 0.787 / 0.752 & 0.975 / 0.824 & 0.839 / 0.785 \\ \hline
        CC   && 0.863 / 0.902 & 0.759 / 0.917 & 0.974 / 0.925 & 0.824 / 0.914 \\ \hline
        VIST  && 0.826 / 0.978 & 0.729 / 0.949 & 0.958 / 0.926 & 0.789 / 0.971 \\ \hline
        COCO  && 0.779 / 1.000 & 0.695 / 1.000 & 0.948 / 1.000 & 0.741 / 1.000 \\ \hline
        SBUCaps,CC  && 0.885 / 0.840 & 0.788 / 0.837 & 0.978 / 0.893 & 0.847 / 0.838  \\ \hline
        R,CC  && 0.876 / 0.888 & 0.801 / 0.784 & 0.976 / 0.918 & 0.855 / 0.852  \\ \hline
        SBUCaps,R & & 0.876 / 0.779 & 0.789 / 0.697 & 0.976 / 0.791 & 0.849 / 0.758  \\ \hline
        SBUCaps &\checkmark  & 0.885 / 0.798 & 0.817 / 0.719 & 0.977 / 0.745 & 0.866 / 0.768  \\ \hline
        R & \checkmark & 0.880 / 0.744 & 0.809 / 0.697 & 0.976 / 0.776 & 0.856 / 0.721 \\ \hline
        CC& \checkmark & 0.868 / 0.913 & 0.765 / 0.920 & 0.975 / 0.942 & 0.835 / 0.920 \\ \hline
        SBUCaps,CC & \checkmark & 0.870 / 0.915 & 0.776 / 0.881 & 0.976 / 0.932 & 0.830 / 0.905  \\ \hline
        R,CC & \checkmark & 0.862 / 0.922 & 0.779 / 0.842 & 0.971 / 0.944 & 0.837 / 0.894 \\ \hline
        SBUCaps,R & \checkmark & 0.877 / 0.807 & 0.805 / 0.712 & 0.973 / 0.856 & 0.844 / 0.828 \\ \hline
        ALL & & 0.860 / 0.969 & 0.779 / 0.903 & 0.973 / 0.990 & 0.832 / 0.947  \\ \hline
    \end{tabular}
        \begin{tabular}{c|c|c|c|c}
    \hline
        \textbf{Train Dataset(s)} & ST  & \textbf{SBUCaps } & \textbf{RedCaps  } & \textbf{CC  } \\ \hline
        No Vetting  &&  0.463 / 1.000 & 0.596 / 1.000 & 0.737 / 1.000 \\ \hline
        SBUCaps   && 0.828 / 0.791 & 0.808 / 0.684 & 0.844 / 0.831 \\ \hline
        RedCaps (R) &  & 0.668 / 0.759 & 0.855 / 0.929 & 0.837 / 0.709 \\ \hline
        CC   &&  0.585 / 0.846 & 0.713 / 0.844 & 0.884 / 0.935 \\ \hline
        VIST  &&  0.518 / 0.939 & 0.658 / 0.883 & 0.771 / 0.981 \\ \hline
        COCO  && 0.463 / 1.000 & 0.599 / 1.000 & 0.739 / 1.000 \\ \hline
        SBUCaps,CC  && 0.923 / 0.950 & 0.762 / 0.822 & 0.965 / 0.978 \\ \hline
        R,CC  && 0.691 / 0.720 & 0.845 / 0.836 & 0.892 / 0.914 \\ \hline
        SBUCaps,R & & 0.892 / 0.940 & 0.923 / 0.958 & 0.846 / 0.785 \\ \hline
        SBUCaps &\checkmark & 0.790 / 0.814 & 0.782 / 0.754 & 0.834 / 0.866 \\ \hline
        R & \checkmark & 0.686 / 0.724 & 0.843 / 0.901 & 0.831 / 0.526 \\ \hline
        CC& \checkmark & 0.609 / 0.841 & 0.721 / 0.862 & 0.922 / 0.955 \\ \hline
        SBUCaps,CC & \checkmark & 0.754 / 0.821 & 0.747 / 0.847 & 0.891 / 0.943 \\ \hline
        R,CC & \checkmark & 0.649 / 0.797 & 0.793 / 0.887 & 0.868 / 0.931 \\ \hline
        SBUCaps,R & \checkmark & 0.826 / 0.724 & 0.804 / 0.905 & 0.839 / 0.771 \\ \hline
        ALL &  & 0.713 / 0.829 & 0.803 / 0.898 & 0.874 / 0.941 \\ \hline
    \end{tabular}
    \caption{Precision and recall of cross-dataset vetting over visual presence validations sets from different sources (DII-VIST...CC). All methods improve precision compared to no vetting.}
    \label{tab:cross_dataset_prec_rec}
\end{table*}
\begin{table*}[!ht]
    \centering
    \begin{tabular}{c|c|c|c|c|c|c|c|c}
    \hline
        \textbf{Train Dataset} & ST & \textbf{DII-VIST } & \textbf{SIS-VIST} & \textbf{COCO } & \textbf{VIST } & \textbf{S } & \textbf{R } & \textbf{CC }\\ \hline
        No Vetting            &   &       0.876         & 0.820 & 0.973                       & 0.851 & 0.633 & 0.747 & 0.849  \\ \hline
        SBUCaps & &0.796 & 0.703 & 0.779                                                            & 0.773 & \cellcolor{lightorange} \textbf{0.809} & 0.741 & 0.837  \\ \hline
        R && 0.828 & 0.769 & 0.893                                                              & 0.811 & \textbf{0.710} & \cellcolor{lightorange} \textbf{0.890} & 0.768  \\ \hline
        CC & &\textbf{0.882} & \textbf{0.830} & 0.949                                                             & \textbf{0.867} & \textbf{0.692} & \textbf{0.773} & \cellcolor{lightorange} \textbf{0.909}  \\ \hline
        VIST & &\cellcolor{lightorange} \textbf{0.895 }& \cellcolor{lightorange} \textbf{0.825} & 0.942           & \cellcolor{lightorange} \textbf{0.871} & \textbf{0.668 }& \textbf{0.754} & \textbf{0.863}  \\ \hline
        COCO && 0.876 & 0.820 & \cellcolor{lightorange}  0.973                                  & 0.851 & 0.633 & \textbf{0.749} & \textbf{0.850}  \\ \hline
        SBUCaps,CC && 0.862 & 0.812 & 0.933                                                     & 0.843 & \cellcolor{lightorange}\textbf{ 0.937} & \textbf{0.791} &\cellcolor{lightorange}  \textbf{0.972}  \\ \hline
        R,CC && \textbf{0.882} & 0.793 & 0.946                                                           & \textbf{0.854} & \textbf{0.705} & \cellcolor{lightorange} \textbf{0.841} & \cellcolor{lightorange} \textbf{0.903}  \\ \hline
        SBUCaps,R & &0.825 & 0.741 & 0.874                                                      & 0.801 &\cellcolor{lightorange} \textbf{ 0.915 }& \cellcolor{lightorange}\textbf{ 0.940} & 0.810  \\ \hline
        SBUCaps &\checkmark& 0.839 & 0.765 & 0.846                                              & 0.814 &\cellcolor{lightorange}\textbf{  0.802} & \textbf{0.767} & \textbf{0.850}  \\ \hline
        R  &\checkmark& 0.806 & 0.749 & 0.865                                                   & 0.783 &\textbf{ 0.705 }&\cellcolor{lightorange} \textbf{ 0.871} & 0.644  \\ \hline
        CC &\checkmark& \textbf{0.890} & \textbf{0.836} & 0.958                                                   & \textbf{0.875} & \textbf{0.707} & \textbf{0.785} & \cellcolor{lightorange} \textbf{0.938}  \\ \hline
        SBUCaps,CC &\checkmark& \textbf{0.892} & \textbf{0.825} & 0.954                                           & \textbf{0.866} &\cellcolor{lightorange}  \textbf{0.786} & \textbf{0.793} & \cellcolor{lightorange} \textbf{0.916 } \\ \hline
        R,CC &\checkmark& \textbf{0.891} & 0.809 & 0.957                                                 & \textbf{0.865} &\textbf{ 0.716} & \cellcolor{lightorange} \textbf{0.837} & \cellcolor{lightorange} \textbf{0.899}  \\ \hline
        SBUCaps,R &\checkmark& 0.841 & 0.756 & 0.911                                            & 0.836 & \cellcolor{lightorange}\textbf{ 0.772} & \cellcolor{lightorange}  \textbf{0.851} & 0.803  \\ \hline
        ALL && \cellcolor{lightorange}  \textbf{0.911} &\cellcolor{lightorange}  \textbf{0.836} &\cellcolor{lightorange}  \textbf{0.981} & \cellcolor{lightorange} \textbf{0.886} &\cellcolor{lightorange}  \textbf{0.767} &\cellcolor{lightorange}  \textbf{0.848} & \cellcolor{lightorange} \textbf{0.906}  \\ \hline
    \end{tabular}
    \caption{F1 scores of cross dataset vetting on visual presence validations sets from different sources (DII-VIST...CC). Datasets abbreviated as S = SBUCaps, R = RedCaps, CC = Conceptual Captions. Bold indicates if result is better than no vetting. Train data containing the same source as the validation is highlighted in yellow.}
    \label{tab:cross_dataset_f1}
\end{table*}

We show results over all the cross-dataset settings we evaluated in Table \ref{tab:cross_dataset_prec_rec}. Notably, this shows that precision in the cross-dataset setting is always better than no vetting except on COCO which already has high precision and differs in composition (more descriptive) compared to the other datasets.

\textbf{Combining multiple datasets.} We find that VEIL is able to leverage additional datasets to an extent. For example, combining SBUCaps and CC leads to significant improvements (7-16\% relative) in F1 as shown in Table \ref{tab:cross_dataset_f1} and, combining SBUCaps and Redcaps in training improves performance on both validation sets. When combining all datasets, only the non-in-the-wild datasets see an improved performance. 

\textbf{Using special token.} We test VEIL$_{\text{ST}}$ which inserts a special token {\tt [EM\_LABEL]} before each extracted label in the caption to reduce the model's reliance on category-specific cues and improve generalization to other datasets. We find that using VEIL w/ ST on average improves F1 by 1 pt compared to just VEIL when transferring to other datasets. This comes at a tradeoff to the performance on the same dataset; however, CC w/ ST improves performance on all datasets.

\begin{table*}[]
    \centering
\begin{tabular}{c|ccc|ccc} \hline
& \multicolumn{3}{c|}{ mAP, IoU }& \multicolumn{3}{c}{mAP, Area} \\\hline
& {0.5:0.95} & 0.5 & 0.75 & S & M & L\\ \hline
GT* & 4.19  & 9.17 & 3.40 & 1.10 & 4.34 & 6.76 \\ \hline
No Vetting & 3.24 & 7.70 & 2.37 & \underline{1.06} & 4.00 & 5.08 \\
Large Loss \cite{Kim2022LargeLM} & 3.11 & 7.54 & 2.15 & 0.92 & 3.80 & 4.88 \\
LocalCLIP-E \cite{Radford2021LearningTV}& 3.66 & 7.77 & 3.08 & 0.79 & 3.96 & 5.96 \\
VEIL$_{\text{ST}}$-R,CC & \underline{3.90} & \underline{8.60} & \underline{3.14} & 0.93 & \underline{4.25} & \underline{6.28} \\
VEIL-SBUCaps & \textbf{4.89} & \textbf{10.37} & \textbf{4.20} & \textbf{1.26} & \textbf{5.24} & \textbf{7.53} \\\hline
\end{tabular}
    \caption{COCO-14 benchmark for WSOD models trained with various vetting methods. (GT*) directly vets labels using the pretrained object detectors which were used to train VEIL. Bold indicates best performance in each column and underline indicates second best result in the column.}
    \label{tab:coco_all_metrics}
\end{table*}

\subsection{WSOD Implementation Details}
\label{appx:WSOD_Implementation_Details}

We used 4 RTX A5000 GPUs and trained for 50k iterations with a batch size of 8, or 100k iterations on 4 Quadro RTX 5000 GPUs with a batch size of 4 and gradient accumulation (parameters updated every two iterations to simulate a batch size of 8).

\textbf{Learning Rates.} We trained four models without vetting on SBUCaps with learning rates from `1e-5' till `1e-2', for each order of magnitude, and observed that the model trained with a learning rate of `1e-2' had substantially better Pascal VOC-07 detection performance. We used this learning rate for all the WSOD models trained on SBUCaps. We applied a similar learning rate selection method for WSOD models trained on RedCaps, except we tested over every half order of magnitude and found that `5e-5' was optimal when training on RedCaps. 

% \subsection{Large Loss Matters Hyperparameter}

\textbf{Relative Delta.} In Large Loss Matters (LLM) \cite{Kim2022LargeLM}, relative delta controls how fast the rejection rate will increase over training. To find the best relative delta, we tested over three initializations, with $rel\_delta=0.002$ as the setting recommended in \cite{Kim2022LargeLM}. We used the best result in Table \ref{tab:rel_delta} when reporting results in the main paper.
\subsection{WSOD Benchmarking on Additional COCO Metrics}
\label{appx:wsod_benchmarking_on_additional_coco_metrics}

In our main text, we compared the average precision of the model across all the classes and all the IoU (Intersection over Union) thresholds from 0.5 to 0.95. We show mAP at specific thresholds 0.5 and 0.75 in Table \ref{tab:coco_all_metrics}. We see that cross-dataset VEIL vetting performs relatively 32\% better than no vetting in a stricter IoU (0.75). The mAP metric can be further broken down by area sizes of ground truth bounding boxes, which is denoted by S, M, and L. VEIL-based vetting outperforms the rest in Medium (6\% better than best non-VEIL vetting) and Large objects (5\% better than best non-VEIL vetting); while VEIL-Same Dataset still performs best on small objects, VEIL-Cross Dataset performs slightly worse than no vetting.

% \subsection{Weighted Sampling Hyperparameter}
% \begin{table}[]
%     \centering
%     \begin{tabular}{c|c}
%         Weighted Sampling & Pascal VOC-07 mAP_{50}  \\
%          0.002 & 28.25 \\
%          0.01 & 30.93\\
%          0.05 & 28.11 \\
%     \end{tabular}
%     \caption{Relative delta hyperparameter ablation}
%     \label{tab:rel_delta}
% \end{table}
% \subsection{Large Loss Matters Hyperparameter}
% \begin{table}[]
%     \centering
%     \begin{tabular}{c|c}
%         Relative Delta & Pascal VOC-07 mAP_{50}  \\
%          0.002 & 28.25 \\
%          0.01 & 30.93\\
%          0.05 & 28.11 \\
%     \end{tabular}
%     \caption{Relative delta hyperparameter ablation}
%     \label{tab:rel_delta}
% \end{table}
\subsection{Additional Qualitative Results}
\label{appx:qual_examples}

\textbf{Vetting Qualitative Examples.} Using annotations from CLaN, we provide qualitative examples comparing the vetting capability of methods on VAELs with common linguistic indicators (prepositional phrase, different word sense, non-literal) found in RedCaps in Figure \ref{fig:vetting_quals}.

\begin{figure*}[t] %[h]
    \centering
\includegraphics[width=\textwidth]{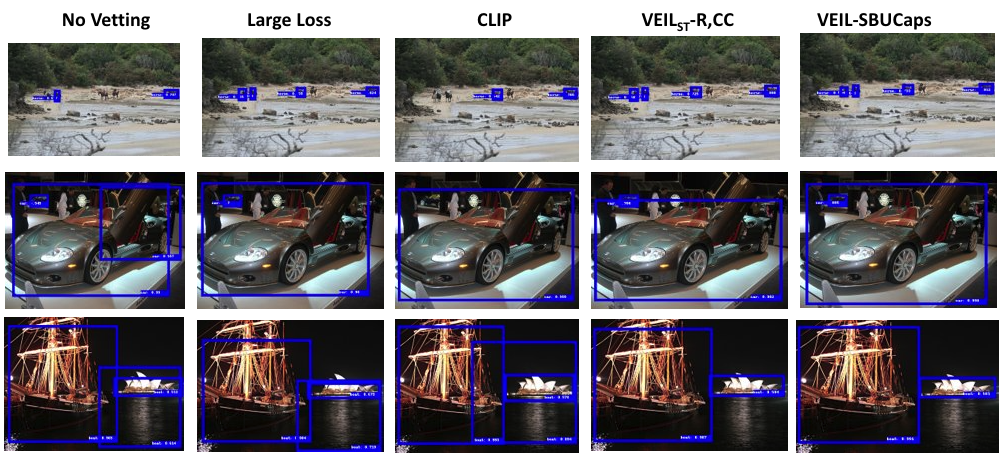}
\includegraphics[width=\textwidth]{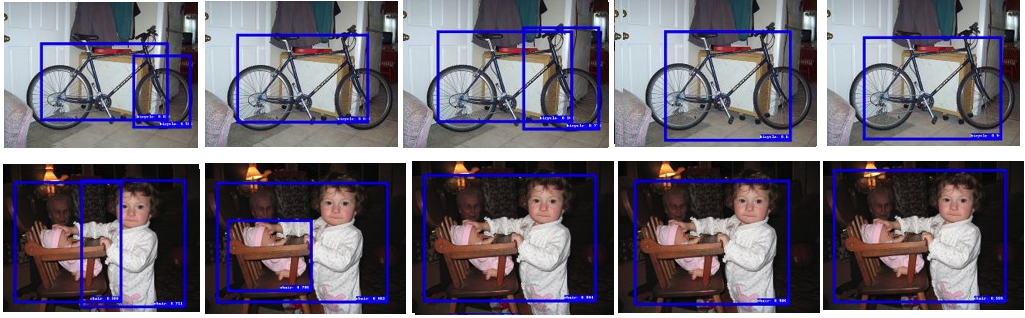}
    %scale=0.8
    \caption{Detections (blue bounding box) from WSOD models trained with various vetting methods (top row) indicate that training with either VEIL-based vetting method (two rightmost columns) leads to similar detection capability on VOC-07 \cite{Everingham2010ThePV}. The categories shown by row (from top to bottom) are: horse, car, boat, bicycle, chair.}
    % adaptive, caption independent approach, LLM \cite{Kim2022LargeLM}.}
    \label{fig:structured_qual}
\end{figure*}

\textbf{WSOD Qualitative Examples.} In Figure \ref{fig:structured_qual}, we present further qualitative evidence on the impact of different vetting methods on weakly supervised object detection. There are varying degrees of part and contextual bias from all methods; however, No Vetting has the most pronounced part domination and context bias as shown by its detection of bicycle wheels and car doors (top two rows), and misidentifying a child as a chair (bottom row) and detections covering both boat and water. Both VEIL methods outperform the rest of the models in detecting smaller objects (see first two rows). LocalCLIP-E misses smaller objects in the background (first two rows) and also has part domination (bicycle).

% This is an appendix.

\end{document}